\documentclass[letterpaper]{article} % DO NOT CHANGE THIS
\usepackage[preprint]{aaai2027}
\usepackage[hyphens]{url}  % DO NOT CHANGE THIS
\usepackage{graphicx} % DO NOT CHANGE THIS
\usepackage{natbib}  % DO NOT CHANGE THIS AND DO NOT ADD ANY OPTIONS TO IT
\usepackage{caption} % DO NOT CHANGE THIS AND DO NOT ADD ANY OPTIONS TO IT
\usepackage{algorithm}
\usepackage{algorithmic}
\usepackage{amsmath}
\usepackage{amssymb}
\usepackage{booktabs}
\usepackage{makecell}

\usepackage{newfloat}
\usepackage{listings}
\usepackage[listings,breakable,skins]{tcolorbox}
\DeclareCaptionStyle{ruled}{labelfont=normalfont,labelsep=colon,strut=off} % DO NOT CHANGE THIS
\floatstyle{ruled}
\newfloat{listing}{tb}{lst}{}
\floatname{listing}{Listing}

\newtcblisting{promptbox}[1]{
  listing only,
  enhanced,
  colback=white,
  colframe=black!65,
  boxrule=0.4pt,
  arc=1pt,
  boxsep=3pt,
  left=3pt,
  right=3pt,
  top=1pt,
  bottom=0pt,
  middle=1pt,
  before skip=2pt,
  after skip=2pt,
  title=\textbf{#1},
  fonttitle=\bfseries,
  listing options={
    basicstyle=\normalfont\footnotesize,
    numbers=none,
    aboveskip=0pt,
    belowskip=0pt,
    breaklines=true,
    breakatwhitespace=false,
    breakautoindent=false,
    breakindent=0pt,
    columns=fullflexible,
    keepspaces=true,
    showstringspaces=false,
    escapeinside={(*@}{@*)}
  }
}

\newcounter{appsubsub}[subsection]
\renewcommand{\theappsubsub}{\Alph{subsection}.\arabic{appsubsub}}
\newcommand{\appthird}[1]{%
  \refstepcounter{appsubsub}%
  \par\medskip\noindent\textbf{\theappsubsub\quad #1}\par\smallskip%
}

\usepackage{booktabs}

\title{LLM-Based Hierarchical Coordinated Control with Continuation-Aware Policy Learning}
\author{
Changhong He\textsuperscript{\rm 1,\rm 2},
Jinda Gao\textsuperscript{\rm 2},
Xinkuan Liu\textsuperscript{\rm 2},
Le Zhang\textsuperscript{\rm 2},
Xizi Luo\textsuperscript{\rm 1,\rm 2},
Yu Mei\textsuperscript{\rm 2}\corresponding
}
\affiliations{
\textsuperscript{\rm 1}Beihang University, Beijing, China\\
\textsuperscript{\rm 2}Baidu, Inc., Beijing, China\\
Corresponding author: whqyqy@hotmail.com
}

\begin{document}

\maketitle

\begin{abstract}
Coordinating multiple interacting units in complex engineering systems is challenging when system interactions are difficult to model, operational information is heterogeneous, and low-level actions must satisfy strict constraints. We propose an LLM-based hierarchical framework in which the LLM coordinates interacting units based on heterogeneous operational context, while task-specific controllers or optimizers generate executable and constraint-aware actions. We further introduce Continuation-Aware GRPO to capture the consequences of coordination decisions over subsequent control intervals. Rather than judging a decision only by its immediate outcome, the method also evaluates how the system evolves afterward under the current policy. We validate the framework on multi-ramp traffic control and virtual power plant (VPP) energy management, using simplified system models for training and more realistic simulators for evaluation. Across both tasks, the proposed method consistently outperforms direct task-specific control and optimization, end-to-end reinforcement learning, rule-based and RL-based hierarchical coordination, and prompting-only LLM coordinators, demonstrating the value of heterogeneous-context reasoning, hierarchical execution, and continuation-aware policy learning.
\end{abstract}

\section{Introduction}
\label{sec:introduction}

Many complex engineering systems consist of multiple interacting units whose local decisions jointly determine system-level performance. Examples include transportation networks, energy systems, industrial processes, and distributed infrastructures, where local actions are coupled through shared dynamics, resource constraints, and global objectives. Effective operation therefore requires not only reliable local controllers, but also an adaptive coordination policy that determines how different units should respond as operating conditions evolve.

% \begin{figure}[!t]
%     \centering
%     \includegraphics[width=\columnwidth]{Figures/introduction_anonymous.pdf}
%     \caption{Introduction of the proposed LLM-based hierarchical coordination
%     framework with continuation-aware policy learning for complex engineering
%     systems. $\checkmark$, $\triangle$, and $\times$ denote full, partial,
%     and no support, respectively.}
%     \label{fig:figure_label}
% \end{figure}

Classical approaches commonly use centralized optimization, decentralized feedback control, or combinations of the two \cite{rawlings2017mpc,negenborn2008multiagent}. Centralized optimization explicitly models system coupling and jointly computes low-level actions, but its effectiveness depends on the accuracy of system dynamics, forecasts, and uncertainty models, while its computational complexity may grow rapidly with system scale. Decentralized feedback control is often more efficient and reliable, but system-level coordination typically relies on fixed parameters or manually designed switching rules \cite{papageorgiou2002ramp,negenborn2008multiagent}. Such rules are difficult to construct when the appropriate strategy depends on temporal trends, spatial interactions, uncertainty, resource urgency, operational priorities, and external events.

Learning-based control reduces the need to specify all coordination rules in advance, but directly learning joint low-level actions remains problematic. The action space grows rapidly with the number of controlled units, physical and operational constraints are difficult to guarantee, and learned policies may overfit the dynamics of their training environments \cite{kiumarsi2018rlcontrol,achiam2017constrained}. Existing LLM-based control approaches face related limitations: prompting alone does not align language-level decisions with closed-loop performance, while directly generating physical actions provides limited guarantees of feasibility and reliability \cite{huang2022zeroshot,huang2023inner,sha2023languagempc}.

We propose an LLM-based hierarchical framework that separates system-level coordination from physical execution. At each decision step, the LLM interprets the current operating context and determines how interacting units should respond—for example, which units should act more aggressively or conservatively and which objectives should be prioritized. Task-specific controllers or optimizers then translate these high-level decisions into executable actions while handling physical and operational constraints.

The LLM is well suited to this role because coordination decisions often depend on heterogeneous information, including numerical measurements, recent trends, uncertainty estimates, resource urgency, operational priorities, external events, and system rules \cite{huang2023inner,sha2023languagempc}. These signals differ in representation, scale, and semantics, yet must be interpreted jointly. Moreover, interactions among controlled units and external disturbances are often difficult to capture with a complete explicit model. Rather than identifying the full system dynamics, the LLM learns a history- and context-conditioned coordination policy from closed-loop experience, while domain-specific controllers retain responsibility for precise execution.

A pretrained LLM, however, is not inherently aligned with control performance. We therefore fine-tune the high-level policy with GRPO using system-level feedback \cite{shao2024deepseekmath}. A key challenge is that the effects of a coordination decision may persist beyond the interval in which it is executed. Short rollouts may therefore favor decisions with immediate benefits but adverse downstream consequences.

To reduce this short-horizon bias, we introduce Continuation-Aware GRPO. Each high-level decision is applied for one control interval, after which the system continues under a frozen copy of the current policy for a longer evaluation horizon. The accumulated return captures both the immediate outcome and the subsequent system evolution, without requiring an additional value model or modifying the standard GRPO objective \cite{shao2024deepseekmath}.

Because the LLM operates at the coordination level rather than generating precise physical actions, its policy is less dependent on fine-grained simulator dynamics. To test whether it learns reusable coordination strategies rather than model-specific heuristics, we train the policy using simplified system models and evaluate it in more realistic environments with different dynamics and modeling assumptions.

We instantiate the framework in two heterogeneous coordinated-control systems: multi-ramp traffic control and virtual power plant energy management \cite{papageorgiou2002ramp,naval2021vpp}. The two case studies differ substantially in physical dynamics, information structure, temporal coupling, constraints, and low-level execution mechanisms. They therefore provide complementary settings for evaluating the generality of the proposed framework.

We evaluate the framework against both direct-control and hierarchical-coordination baselines across diverse operating patterns and simulator dynamics.

Our main contributions are as follows:
\begin{itemize}
\item We propose an LLM-based hierarchical framework for coordinated control, in which the LLM integrates heterogeneous operational context and makes system-level coordination decisions, while task-specific controllers or optimizers provide precise and constraint-aware execution.
\item We introduce a closed-loop post-training approach for the high-level LLM policy. The approach incorporates a continuation-aware return construction into GRPO, evaluating each coordination decision over both its execution interval and the subsequent system evolution under a frozen copy of the current policy.
\item We validate the framework on two structurally different engineering systems and demonstrate consistent improvements over both direct-control and hierarchical-coordination baselines across diverse operating patterns, as well as reduced performance degradation under changes in simulator dynamics.
\end{itemize}

\section{Related Work}

\subsection{Traditional and Learning-Based Control}
\label{sec:traditional_learning_control}

Complex engineering systems commonly use feedback, model-based
optimization, or hybrid control. Traffic applications employ feedback
and model predictive control, while virtual power plants use
optimization to coordinate distributed resources
\cite{papageorgiou2002ramp,hegyi2005mpc,naval2021vpp}.
Feedback control is efficient but often relies on fixed coordination
logic, whereas centralized optimization captures system couplings and
constraints but depends on accurate models and forecasts and can be
costly at scale.

Reinforcement learning reduces reliance on predefined rules, but direct
low-level policies must learn coordination, numerical actions, and
constraint satisfaction jointly
\cite{belletti2018rampdrl,liu2023vpprl}. Hierarchical methods instead
combine learned high-level decisions with domain-specific controllers
or optimizers \cite{airaldi2025rlmpc,li2025hierarchicalvpp}, but their
representations, coordination variables, and execution interfaces
remain task- and model-specific.

\subsection{LLM-Based Decision Making and Control}
\label{sec:llm_decision_control}

LLMs have been used in engineering control as action selectors,
supervisors, and interfaces to numerical controllers. LLMLight and
CoLLMLight map textual traffic states to signal phases, with CoLLMLight
additionally incorporating neighboring states and historical evolution
\cite{lai2025llmlight,yuan2026collmlight}. Other approaches couple LLMs
with model predictive control, reinforcement learning, or optimization
through high-level decisions, predictions, parameter adjustment, or
formulation generation
\cite{sha2023languagempc,jin2024llmenergy,wu2025instructmpc}.
LLMASC uses language-based proposal exchange and consensus before
low-level RL execution \cite{yu2026llmasc}. Despite their flexibility,
these approaches define LLM outputs and numerical execution interfaces
separately for each application.

\subsection{Reinforcement Fine-Tuning of LLMs}
\label{sec:reinforcement_finetuning}

Reinforcement fine-tuning aligns LLM policies with task feedback.
DeepSeekMath introduced Group Relative Policy Optimization (GRPO),
which estimates relative advantages among outputs sampled for the same
prompt without a value model, while DeepSeek-R1 demonstrated scalable
reinforcement learning with verifiable rewards
\cite{shao2024deepseekmath,guo2025deepseekr1}. These methods primarily
evaluate complete outputs in static domains such as mathematics and
code.

Recent work extends reinforcement fine-tuning to stateful interaction:
LOOP learns from complete environment trajectories, turn-level methods
improve credit assignment in multi-turn tool use, and Traffic-R1 uses
simulator feedback for sequential traffic control
\cite{chen2025loop,zeng2025turncredit,zou2025trafficr1}.

Engineering control poses a distinct temporal mismatch: a decision
executed during one interval changes physical states and may affect
performance over subsequent intervals. Existing formulations do not
directly specify how to evaluate that decision through a longer
closed-loop continuation, motivating our continuation-aware return
construction.

\section{Problem Formulation}
\label{sec:problem}

We consider a complex engineering system composed of $N$ interacting
control units. At time $t$, the physical state $s_t \in \mathcal{S}$
contains the observable system variables, and the executable action
$\mathbf{a}_t \in \mathcal{A}$ affects the system through
\begin{equation}
s_{t+1}
=
f(s_t,\mathbf{a}_t,w_t),
\label{eq:system_dynamics}
\end{equation}
where $w_t$ denotes exogenous disturbances such as time-varying demand,
renewable generation, weather, or market conditions. The executable
action must satisfy the operational constraints represented in the
control model,
\begin{equation}
g(s_t,\mathbf{a}_t) \leq 0.
\label{eq:constraints}
\end{equation}

In addition to the instantaneous physical state, coordinated decisions
may depend on recent system evolution and other operational information.
We denote the available decision context by
\begin{equation}
\xi_t = (s_{t-L:t},c_t),
\label{eq:operational_context}
\end{equation}
where $s_{t-L:t}$ is the recent state history and $c_t$ contains
additional information such as forecasts, uncertainty estimates,
resource urgency, operational priorities, external events, and system
rules.

Rather than directly generating the physical action, the high-level
policy produces a structured joint coordination decision
\begin{equation}
\mathbf{z}_t
=
\left(z_t^1,\ldots,z_t^N\right)
\in
\mathcal{Z}
=
\mathcal{Z}_1 \times \cdots \times \mathcal{Z}_N,
\label{eq:joint_decision}
\end{equation}
where $z_t^n$ specifies how control unit $n$ should respond under the
current operating context. A task-specific low-level controller or
optimizer converts this joint decision into an executable action,
\begin{equation}
\mathbf{a}_t
=
\pi_{\mathrm{low}}(s_t;\mathbf{z}_t).
\label{eq:hierarchical_execution}
\end{equation}

The high-level coordination policy is parameterized by an LLM,
\begin{equation}
\mathbf{z}_t
\sim
\pi_\phi(\cdot \mid \xi_t).
\label{eq:high_level_policy}
\end{equation}
For a trajectory
$\tau=(s_0,\mathbf{a}_0,w_0,\ldots,s_T)$, system-level performance is
measured by
\begin{equation}
R(\tau)
=
\sum_{t=0}^{T-1}
\gamma^t
r(s_t,\mathbf{a}_t,w_t),
\label{eq:trajectory_return}
\end{equation}
where $\gamma \in (0,1]$ is the discount factor.

The learning objective is to optimize the high-level policy while the
low-level controller remains responsible for physical execution:
\begin{equation}
\begin{aligned}
\max_{\phi}\quad
&
\mathbb{E}_{\tau \sim p_\phi}
\left[
R(\tau)
\right]
\\
\text{s.t.}\quad
&
\mathbf{z}_t
\sim
\pi_\phi(\cdot \mid \xi_t),
\\
&
\mathbf{a}_t
=
\pi_{\mathrm{low}}(s_t;\mathbf{z}_t),
\\
&
g(s_t,\mathbf{a}_t) \leq 0,
\qquad
t=0,\ldots,T-1,
\end{aligned}
\label{eq:objective}
\end{equation}
where $p_\phi$ is the trajectory distribution induced by the high-level
policy, the low-level controller, the system dynamics, and the
exogenous disturbances.

\section{Method}
\label{sec:method}

\begin{figure*}[!t]
    \centering
    \includegraphics[width=\textwidth]{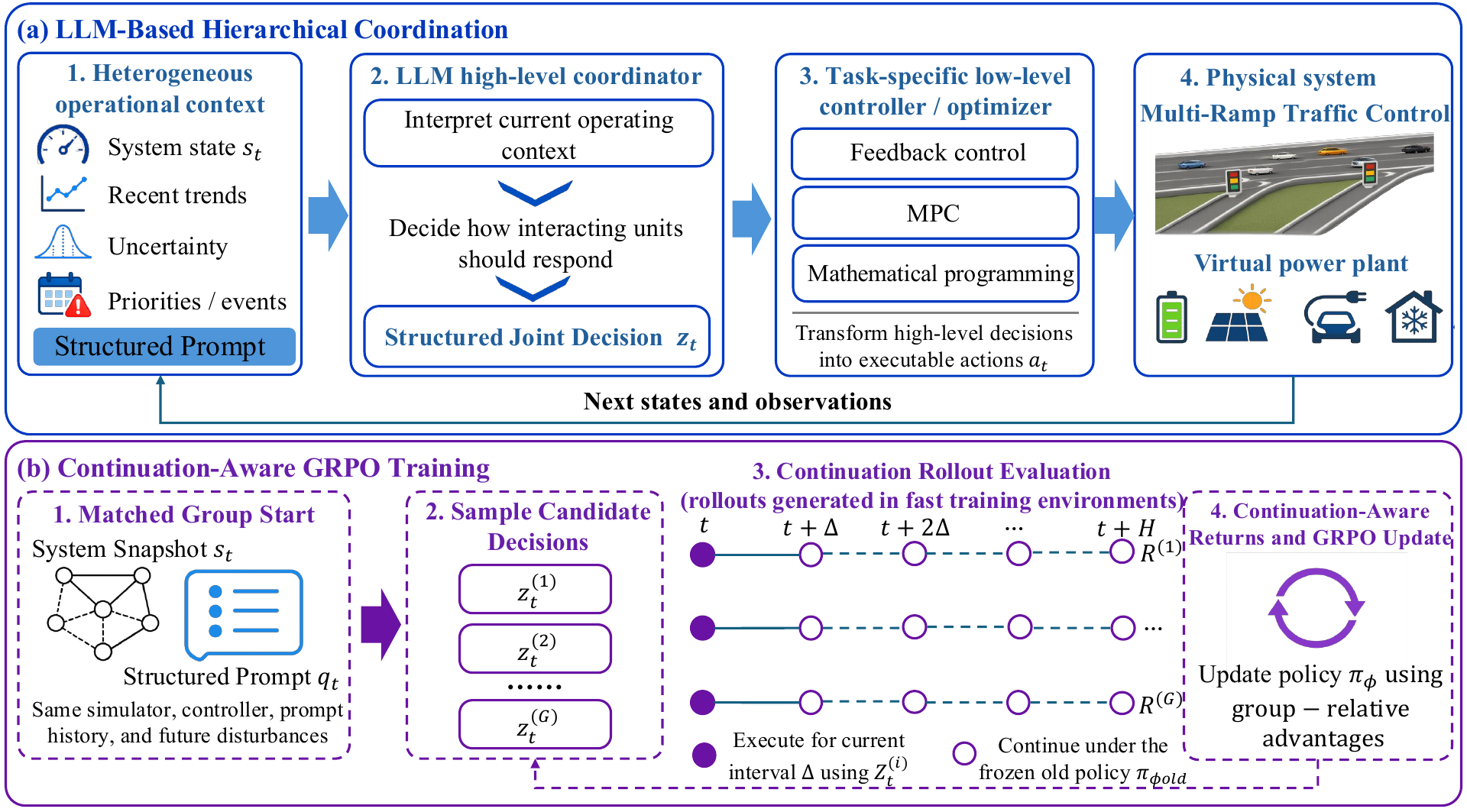}
    \caption{Overview of the proposed LLM-based hierarchical coordination framework and Continuation-Aware GRPO.}
    \label{fig:framework}
\end{figure*}

Figure~\ref{fig:framework}(a) illustrates the hierarchical coordination
architecture, while Figure~\ref{fig:framework}(b) presents the
Continuation-Aware GRPO training procedure.

\subsection{Operational-Context Representation and Hierarchical Execution}
\label{sec:method_architecture}

\paragraph{Structured operational context.}
The heterogeneous context $\xi_t$ is converted into a structured
natural-language prompt
\begin{equation}
q_t
=
\operatorname{Prompt}(\xi_t).
\label{eq:prompt}
\end{equation}
The prompt organizes the available information into task-relevant
descriptions of the current system condition, recent trends,
uncertainty, urgency, priorities, external events, and applicable
rules. 

The LLM policy generates the structured joint decision defined in
Equation~\ref{eq:joint_decision},
\begin{equation}
\mathbf{z}_t
\sim
\pi_\phi(\cdot \mid q_t).
\label{eq:llm_policy}
\end{equation}
During training, decisions are sampled to support policy exploration.
During evaluation, the policy uses a fixed decoding rule.

\paragraph{High-level to low-level interface.}
Each component $z_t^n$ is mapped to parameters used by the corresponding
low-level controller. We write this mapping as
\begin{equation}
\Theta(\mathbf{z}_t)
=
\left[
\theta_1(z_t^1),\ldots,\theta_N(z_t^N)
\right],
\label{eq:parameter_mapping}
\end{equation}
where $\theta_n(z_t^n)$ may represent a feedback threshold, an objective
weight, a resource priority, or another task-specific control
parameter.

The executable action is then computed as
\begin{equation}
\mathbf{a}_t
=
\pi_{\mathrm{low}}
\left(
s_t;
\Theta(\mathbf{z}_t)
\right).
\label{eq:low_level_control}
\end{equation}
The low-level component may use feedback control, model predictive
control, or mathematical programming. It performs the numerical
computation and handles the physical and operational constraints
explicitly encoded in the task model.

The two layers therefore have complementary responsibilities. The LLM
interprets heterogeneous operational information and determines the
system-level coordination strategy. The low-level controller converts
that strategy into executable and constraint-aware physical actions.

Task-specific system models, controller implementations, parameter
mappings, prompt templates, and output schemas are provided in
Appendices~A and~B.

\subsection{Continuation-Aware GRPO}
\label{sec:method_grpo}

A pretrained LLM is not inherently optimized for closed-loop system
performance. We therefore fine-tune the high-level policy using
system-level returns. Standard short-rollout evaluation, however, can
be overly myopic because the effects of a coordination decision may
persist after the interval in which it is executed.

Continuation-Aware GRPO separates the duration of the current decision
from the horizon used to evaluate its consequences.

\paragraph{Matched group evaluation.}
Training samples are constructed from system snapshots extracted at
different points of simulated trajectories. For a snapshot with context
$q_t$, the frozen old policy generates a group of $G$ joint decisions,
\begin{equation}
\left\{
\mathbf{z}_t^{(i)}
\right\}_{i=1}^{G}
\sim
\pi_{\phi_{\mathrm{old}}}(\cdot \mid q_t).
\label{eq:group_sampling}
\end{equation}

All group members start from the same physical state and share the same
future exogenous trajectory. They differ only in the high-level
decision applied during the first control interval. This matched
evaluation ensures that differences in their returns are attributable
primarily to the coordination decisions rather than to different
realizations of demand, renewable generation, weather, prices, or other
external conditions.

\paragraph{One-interval decision and longer continuation.}
Each sampled decision $\mathbf{z}_t^{(i)}$ is applied for one high-level
control interval of duration $\Delta$. The resulting low-level actions
are computed using Equation~\ref{eq:low_level_control}.

After this interval, each rollout continues until the evaluation horizon
$H$, where $H>\Delta$. Subsequent high-level decisions are generated by
the same frozen old policy,
\begin{equation}
\mathbf{z}_{\tau}^{(i)}
\sim
\pi_{\phi_{\mathrm{old}}}
\left(
\cdot \mid q_{\tau}^{(i)}
\right),
\
\tau=t+\Delta,t+2\Delta,\ldots,t+H-\Delta.
\label{eq:continuation_policy}
\end{equation}
The policy parameters and decoding rule remain fixed throughout the
continuation.

The continuation policy is shared at the policy level rather than at
the action-sequence level. Since the initial decisions produce
different subsequent states, the frozen policy may select different
follow-up decisions in different rollouts. This preserves closed-loop
adaptation while evaluating every initial decision under the same
subsequent policy.

The return assigned to the initial decision is
\begin{equation}
R^{(i)}
=
\sum_{k=0}^{H-1}
\gamma^k
r
\left(
s_{t+k}^{(i)},
\mathbf{a}_{t+k}^{(i)},
w_{t+k}
\right).
\label{eq:continuation_return}
\end{equation}
No additional terminal-value model or auxiliary reward is introduced.
When $H=\Delta$, this evaluation reduces to the short-rollout setting.
For $H>\Delta$, it captures both the immediate outcome and the
subsequent system evolution caused by the initial coordination
decision.

\paragraph{Group-relative policy update.}
The group returns are normalized to obtain the relative advantages
\begin{equation}
\widehat{A}^{(i)}
=
\frac{
R^{(i)}-\mu_R
}{
\sigma_R+\varepsilon
},
\label{eq:advantage}
\end{equation}
where $\mu_R$ and $\sigma_R$ are the mean and standard deviation of
returns within the group.

The importance ratio is defined as
\begin{equation}
\rho^{(i)}(\phi)
=
\frac{
\pi_\phi(\mathbf{z}_t^{(i)} \mid q_t)
}{
\pi_{\phi_{\mathrm{old}}}
(\mathbf{z}_t^{(i)} \mid q_t)
}.
\label{eq:importance_ratio}
\end{equation}
We further define the clipped ratio
\begin{equation}
\widetilde{\rho}^{(i)}(\phi)
=
\operatorname{clip}
\left(
\rho^{(i)}(\phi),
1-\epsilon_c,
1+\epsilon_c
\right).
\label{eq:clipped_ratio}
\end{equation}

The policy is optimized using
\begin{equation}
\begin{aligned}
\mathcal{L}(\phi)
={}&
-\mathbb{E}
\left[
\frac{1}{G}
\sum_{i=1}^{G}
\min
\left(
\rho^{(i)}(\phi)\widehat{A}^{(i)},
\widetilde{\rho}^{(i)}(\phi)\widehat{A}^{(i)}
\right)
\right]
\\
&+
\beta
D_{\mathrm{KL}}
\left(
\pi_\phi(\cdot \mid q_t)
\,\|\, 
\pi_{\phi_{\mathrm{ref}}}(\cdot \mid q_t)
\right).
\end{aligned}
\label{eq:grpo}
\end{equation}
Here, $\epsilon_c$ is the clipping threshold,
$\pi_{\phi_{\mathrm{ref}}}$ is the reference policy, and
$\beta$ controls KL regularization.

Continuation-Aware GRPO retains the standard group-relative policy
update. Its distinguishing feature is the return construction. Each
initial decision is executed for one control interval but evaluated
through a longer closed-loop trajectory.

Task-specific reward construction and additional training details are
provided in Appendix~B.

\subsection{Training and Evaluation Across Operating Patterns and Simulators}

During training, the policy interacts with a computationally efficient
simulator $\mathcal{E}_{\mathrm{train}}$ that preserves the dominant
dynamics and interactions required for high-level coordination.
Training scenarios span multiple operating-pattern families and provide
the system snapshots and closed-loop trajectories used by
Continuation-Aware GRPO.

During evaluation, the frozen policy is tested on new instances of the
training pattern families, previously unseen patterns, and scenarios
with increased uncertainty. It is also evaluated without additional
fine-tuning in a more realistic simulator $\mathcal{E}_{\mathrm{eval}}$
with different dynamics and modeling assumptions. The
operational-context format, high-level decision space, and low-level
control interface remain unchanged across simulators. These settings
assess robustness to changes in both operating conditions and simulator
dynamics. Task-specific settings are described below; additional simulator
details, interface correspondences, and modeling differences are
provided in Appendix~C.

\section{Experiments}
\label{sec:experiments}
\begin{table*}[!t]
\centering
\small
\setlength{\tabcolsep}{4.5pt}
\begin{tabular}{@{}lrrrrr@{}}
\toprule
Method & Seen & Unseen & High-U & Overall & Gain \\
\midrule

\multicolumn{6}{l}{\textit{Direct Control}} \\

Feedback Control
    & 16283.4 \(\pm\) 41.8
    & 15576.8 \(\pm\) 50.6
    & 14917.3 \(\pm\) 68.3
    & 15592.5 \(\pm\) 49.1
    & -- \\

End-to-End RL
    & 16537.6 \(\pm\) 163.7
    & 15684.9 \(\pm\) 191.5
    & 14813.2 \(\pm\) 228.4
    & 15678.6 \(\pm\) 181.2
    & +0.55\% \\

\midrule
\multicolumn{6}{l}{\textit{Hierarchical Coordination}} \\

Fixed-Mode Hierarchy
    & 16502.8 \(\pm\) 36.2
    & 15792.4 \(\pm\) 44.9
    & 15158.7 \(\pm\) 60.5
    & 15818.0 \(\pm\) 43.7
    & +1.45\% \\

Hierarchical RL
    & 16843.1 \(\pm\) 98.6
    & 16117.6 \(\pm\) 121.4
    & 15434.8 \(\pm\) 151.8
    & 16131.8 \(\pm\) 110.2
    & +3.46\% \\

Qwen3-8B
    & 16283.4 \(\pm\) 45.7
    & 15177.8 \(\pm\) 59.8
    & 14350.9 \(\pm\) 78.6
    & 15270.7 \(\pm\) 55.1
    & -2.06\% \\

Gemini 3.1 Pro
    & 16422.6 \(\pm\) 43.1
    & 15664.3 \(\pm\) 55.2
    & 14892.1 \(\pm\) 72.7
    & 15659.7 \(\pm\) 50.8
    & +0.43\% \\

Claude Sonnet 4.5
    & 16378.4 \(\pm\) 44.8
    & 15612.7 \(\pm\) 57.6
    & 14847.9 \(\pm\) 75.4
    & 15613.0 \(\pm\) 52.7
    & +0.13\% \\

\textbf{Ours}
    & \textbf{17043.6 \(\pm\) 75.9}
    & \textbf{16617.9 \(\pm\) 92.7}
    & \textbf{16214.7 \(\pm\) 118.6}
    & \textbf{16625.4 \(\pm\) 84.5}
    & \textbf{+6.62\%} \\

\bottomrule
\end{tabular}

\caption{Traffic throughput in SUMO over five matched runs
(mean \(\pm\) SD). Gain is relative to Feedback Control;
higher is better.}
\label{tab:ramp_main}
\end{table*}

\begin{table*}[!t]
\centering
\small
\setlength{\tabcolsep}{4.5pt}
\renewcommand{\arraystretch}{1.08}
\begin{tabular}{@{}lrrrrr@{}}
\toprule
Method & Seen & Unseen & High-U & Overall & Gain \\
\midrule

\multicolumn{6}{l}{\textit{Direct Control}} \\

MPC-MILP
    & 32783.5 \(\pm\) 47.2
    & 33887.9 \(\pm\) 65.8
    & 35843.3 \(\pm\) 111.4
    & 34171.6 \(\pm\) 64.2
    & -- \\

End-to-End RL
    & 35076.9 \(\pm\) 512.4
    & 37423.6 \(\pm\) 647.3
    & 40947.2 \(\pm\) 923.8
    & 37815.9 \(\pm\) 608.5
    & \(-10.66\%\) \\

\midrule
\multicolumn{6}{l}{\textit{Hierarchical Coordination}} \\

Fixed-Mode Hierarchy
    & 32984.7 \(\pm\) 52.6
    & 34052.2 \(\pm\) 71.9
    & 35796.4 \(\pm\) 119.8
    & 34277.8 \(\pm\) 70.6
    & \(-0.31\%\) \\

Hierarchical RL
    & 32183.7 \(\pm\) 267.5
    & 33237.2 \(\pm\) 351.6
    & 35124.6 \(\pm\) 505.2
    & 33515.2 \(\pm\) 317.4
    & +1.92\% \\

Qwen3-8B
    & 34821.7 \(\pm\) 115.8
    & 36542.4 \(\pm\) 158.4
    & 39264.6 \(\pm\) 251.6
    & 36876.2 \(\pm\) 144.7
    & \(-7.91\%\) \\

Gemini 3.1 Pro
    & 33486.5 \(\pm\) 106.3
    & 34726.3 \(\pm\) 144.9
    & 37218.5 \(\pm\) 231.7
    & 35143.8 \(\pm\) 132.5
    & \(-2.85\%\) \\

Claude Sonnet 4.5
    & 33634.2 \(\pm\) 110.7
    & 34918.7 \(\pm\) 150.6
    & 37406.8 \(\pm\) 240.8
    & 35319.9 \(\pm\) 138.3
    & \(-3.36\%\) \\

\textbf{Ours}
    & \textbf{31937.9 \(\pm\) 183.6}
    & \textbf{32583.5 \(\pm\) 246.8}
    & \textbf{33516.3 \(\pm\) 371.5}
    & \textbf{32679.2 \(\pm\) 223.6}
    & \textbf{+4.37\%} \\

\bottomrule
\end{tabular}

\caption{VPP operating cost in OpenDSS over five matched runs
(mean \(\pm\) SD). Gain is relative to MPC-MILP;
lower is better.}
\label{tab:vpp_main}
\end{table*}

We evaluate the proposed framework on two structurally distinct
coordinated-control systems: multi-ramp traffic control and virtual
power plant energy management. The experiments address three questions:
(i) how well the framework performs across seen-pattern, unseen-pattern,
and higher-uncertainty scenarios; (ii) whether it remains effective
without target-environment adaptation in environments with more realistic
dynamics and different modeling assumptions; and (iii) how hierarchical execution and continuation-aware return
construction contribute to closed-loop performance.

\subsection{Experimental Tasks and Settings}
\label{sec:experimental_tasks_data}

For each task, we train a separate high-level policy using 150 training
and 50 validation scenarios. We evaluate the frozen policy without
target-environment adaptation on 300 test scenarios, divided equally
into disjoint seen-pattern, unseen-pattern, and higher-uncertainty
subsets. Complete-scenario splits prevent trajectory leakage. Further
details are provided in Appendix~C.

\textbf{Multi-ramp traffic control.}
The traffic task considers eight controlled on-ramps on the northbound
Western Expressway in Changchun~\cite{gao2026dynamicramp}. Each scenario
comprises a 15-minute warm-up and a 3-hour evaluation period. Low-level
hysteresis controllers operate every minute, while the high-level policy
selects joint ramp modes every 3 minutes. Demand trajectories vary
temporal profiles, ramp-wise distributions, and stochastic
perturbations. Training uses the link transmission
model (LTM)~\cite{yperman2005ltm}; evaluation uses
SUMO~\cite{lopez2018sumo}, which represents vehicle-level dynamics, lane
changing, and merging. Seen-pattern scenarios draw new trajectories
from the training pattern family; unseen-pattern scenarios introduce
shifted and double peaks; higher-uncertainty scenarios increase demand
perturbations and forecast errors.

\textbf{Virtual power plant energy management.}
The virtual power plant aggregates photovoltaic generation, battery
storage, electric vehicles, and heating, ventilation, and air
conditioning. Photovoltaic and temperature trajectories use NASA POWER
data~\cite{nasaPower2020}, loads use Building Data Genome
2~\cite{miller2020building}, and electric-vehicle availability uses
ACN-Data~\cite{lee2019acndata}; prices follow perturbed time-of-use
profiles. Each scenario covers 10:00--22:00. The low-level mixed-integer
linear program operates every 5 minutes, while the high-level policy
selects resource modes every 30 minutes. Training uses the fast
single-bus model; evaluation uses the EPRI Ckt5 feeder in
OpenDSS~\cite{dugan2011opendss,epriCkt5}, incorporating feeder-level
power flow, voltage variation, losses, and network constraints.
Seen-pattern scenarios use new trajectories with familiar operating
combinations; unseen-pattern scenarios use withheld combinations;
higher-uncertainty scenarios increase forecast errors and
electric-vehicle timing uncertainty.

\subsection{Baselines and Evaluation Protocol}
\label{sec:baselines_evaluation}

\textbf{Baselines.}
We organize the baselines according to whether they generate executable low-level actions directly or make high-level coordination decisions that are subsequently executed by task-specific low-level components.

The \emph{direct-control baselines} include a task-specific conventional method and an end-to-end reinforcement learning policy. For multi-ramp traffic control, the conventional baseline is the published feedback controller~\cite{gao2026dynamicramp}; for virtual power plant management, it is an MPC-MILP method with fixed configurations~\cite{zapata2014vpp}. The end-to-end RL baseline directly generates low-level control actions using PPO~\cite{schulman2017proximal}.

The \emph{hierarchical-coordination baselines} all operate over the same joint mode space and use the same task-specific low-level controllers or optimizers as our method. Fixed-Mode Hierarchy selects a single joint mode configuration based on validation performance and applies it throughout evaluation. Hierarchical RL is a learned non-LLM coordinator that maps an information-equivalent structured representation of the operational context to joint modes using PPO. The prompting-only LLM coordinators, Qwen3-8B~\cite{yang2025qwen3}, Gemini 3.1 Pro~\cite{google2026gemini31pro}, and Claude Sonnet 4.5~\cite{anthropic2025claudesonnet45}, receive the same operational information as our policy and select joint modes through the same structured output interface, but receive no task-specific parameter updates.

For each task, our policy is initialized from Qwen3-8B and fine-tuned using only Continuation-Aware GRPO. All trainable methods use the same training and validation scenarios, and model selection is based solely on validation performance. Implementation details are provided in Appendix~B.

\begin{table*}[!t]
\centering
\footnotesize
\setlength{\tabcolsep}{3pt}
\renewcommand{\arraystretch}{1.10}
\begin{tabular*}{\textwidth}
{@{\extracolsep{\fill}}lrrrrrr@{}}
\toprule
& \multicolumn{3}{c}{Multi-Ramp Traffic Control}
& \multicolumn{3}{c}{Virtual Power Plant} \\
\cmidrule(lr){2-4}
\cmidrule(lr){5-7}

Method
& LTM \(\uparrow\)
& SUMO \(\uparrow\)
& Gap \(\downarrow\)
& Fast Simulator \(\downarrow\)
& OpenDSS \(\downarrow\)
& Gap \(\downarrow\) \\
\midrule

Qwen3-8B
    & 16312.6 \(\pm\) 33.4
    & 15270.7 \(\pm\) 55.1
    & 6.39 \(\pm\) 0.21\%
    & 34025.2 \(\pm\) 87.9
    & 36876.2 \(\pm\) 144.7
    & 8.38 \(\pm\) 0.26\% \\

End-to-End RL
    & \textbf{17438.7 \(\pm\) 147.8}
    & 15678.6 \(\pm\) 181.2
    & 10.09 \(\pm\) 1.07\%
    & 33785.3 \(\pm\) 438.6
    & 37815.9 \(\pm\) 608.5
    & 11.94 \(\pm\) 2.03\% \\

Hierarchical RL
    & 17043.9 \(\pm\) 91.5
    & 16131.8 \(\pm\) 110.2
    & 5.35 \(\pm\) 0.54\%
    & \textbf{30684.7 \(\pm\) 229.4}
    & 33515.2 \(\pm\) 317.4
    & 9.23 \(\pm\) 0.95\% \\

\textbf{Ours}
    & 17286.4 \(\pm\) 66.7
    & \textbf{16625.4 \(\pm\) 84.5}
    & \textbf{3.82 \(\pm\) 0.33\%}
    & 30941.8 \(\pm\) 142.7
    & \textbf{32679.2 \(\pm\) 223.6}
    & \textbf{5.62 \(\pm\) 0.54\%} \\

\bottomrule
\end{tabular*}

\caption{Cross-simulator performance over five matched runs
(mean \(\pm\) SD). Gap measures degradation from the fast
to the more realistic simulator.}
\label{tab:cross_simulator}
\end{table*}

\begin{table*}[!t]
\centering
\small
\setlength{\tabcolsep}{4.5pt}
\renewcommand{\arraystretch}{1.08}

\textbf{(a) Multi-Ramp Traffic Control: Throughput \(\uparrow\)}
\vspace{3pt}

\begin{tabular}{@{}lrrrrr@{}}
\toprule
Variant & Seen & Unseen & High-U & Overall & Gain \\
\midrule

GRPO (\(H=\Delta\))
    & 16861.4 \(\pm\) 112.8
    & 16327.6 \(\pm\) 151.3
    & 15767.1 \(\pm\) 211.6
    & 16318.7 \(\pm\) 136.4
    & -- \\

\textbf{Ours} (\(H=4\Delta\))
    & \textbf{17043.6 \(\pm\) 75.9}
    & \textbf{16617.9 \(\pm\) 92.7}
    & \textbf{16214.7 \(\pm\) 118.6}
    & \textbf{16625.4 \(\pm\) 84.5}
    & \textbf{+1.88\%} \\

\bottomrule
\end{tabular}

\textbf{(b) Virtual Power Plant: Operating Cost \(\downarrow\)}
\vspace{3pt}

\begin{tabular}{@{}lrrrrr@{}}
\toprule
Variant & Seen & Unseen & High-U & Overall & Gain \\
\midrule

GRPO (\(H=\Delta\))
    & 32463.7 \(\pm\) 264.7
    & 33437.6 \(\pm\) 376.5
    & 34843.1 \(\pm\) 589.2
    & 33581.5 \(\pm\) 327.8
    & -- \\

\textbf{Ours} (\(H=4\Delta\))
    & \textbf{31937.9 \(\pm\) 183.6}
    & \textbf{32583.5 \(\pm\) 246.8}
    & \textbf{33516.3 \(\pm\) 371.5}
    & \textbf{32679.2 \(\pm\) 223.6}
    & \textbf{+2.69\%} \\

\bottomrule
\end{tabular}

\caption{Ablation of continuation-aware return construction over five runs with matched evaluation seeds (mean \(\pm\) SD). Gain is relative to
GRPO with \(H=\Delta\).}
\label{tab:continuation_ablation}
\end{table*}

\textbf{Evaluation protocol.}
We repeat each experiment five times. Trainable methods use independent
training seeds, and all methods use matched simulator seeds. Within each
run, all methods share the same initial state, exogenous inputs,
stochastic realization, and simulation seed. All LLM policies use
deterministic decoding at temperature zero. The primary metrics are
corridor throughput for traffic control, where higher is better, and
realized operating cost for VPP management, where lower is better. For each run, we average scenarios within each subset and average the
three subset scores equally to obtain the overall result. Tables report
the mean \(\pm\) standard deviation of these run-level scores. Gains are
measured against Feedback Control for traffic and MPC-MILP for VPP,
with throughput increases and cost reductions reported as positive.
Evaluation details are provided in Appendix C.

\subsection{Main Results}
\label{sec:main_results}

Tables 1 and 2 report performance in
SUMO and OpenDSS, respectively. Our method obtains the highest mean
throughput and the lowest mean operating cost in every test subset.
Across five runs, it achieves an overall traffic throughput of
\(16625.4 \pm 84.5\), improving over Feedback Control by 6.62\% and
Hierarchical RL by 3.06\%. For VPP management, it achieves an operating
cost of \(32679.2 \pm 223.6\), reducing cost by 4.37\% relative to
MPC-MILP and by 2.49\% relative to Hierarchical RL.

Comparisons across baseline families provide complementary evidence.
Our method outperforms the fixed-mode hierarchy in both tasks, consistent
with the benefit of adapting joint modes to the current operational
context. Hierarchical RL also outperforms end-to-end RL
under the shared evaluation protocol, consistent with retaining
task-specific low-level execution. Our method further improves over
hierarchical RL while using the same joint mode space and
task-specific low-level components. Prompting-only Qwen3-8B, Gemini 3.1
Pro, and Claude Sonnet 4.5 do not match the learned policy. Under this
protocol, prompting alone does not attain the performance of
task-specific learning from closed-loop feedback.

Based on the subset means, our method also shows the smallest observed
degradation from the seen-pattern to the higher-uncertainty subset. Its traffic throughput
decreases by \(4.86\%\), compared with \(8.14\%\)--\(11.87\%\) for the
baselines, while its virtual power plant operating cost increases by
\(4.94\%\), compared with \(8.52\%\)--\(16.74\%\). Its leading mean
performance on both unseen-pattern and higher-uncertainty scenarios
suggests that the learned coordination policy remains effective as
operating patterns and uncertainty levels change. Additional metrics
and results are provided in Appendix~D.

\subsection{Cross-Simulator Evaluation}

Trainable policies are trained only in the fast simulator and evaluated
with frozen parameters, without target-environment adaptation, in a more
realistic simulator with different dynamics and modeling assumptions.
The prompting-only configuration is unchanged. The simulator pairs are
LTM--SUMO for traffic and single-bus--OpenDSS for virtual power plant
management. Within each pair, evaluations use matched initial conditions
and exogenous trajectories, together with the same metric definitions
and evaluation horizons. For mode-level methods, the operational-context
format, joint mode space, and low-level control interface also remain
unchanged.

Table~\ref{tab:cross_simulator} reports mean \(\pm\) standard deviation
over five paired runs, with each gap computed before aggregation. For
traffic, the gap is the throughput decrease from LTM to SUMO relative
to LTM throughput; for VPP management, it is the operating-cost increase
from the fast simulator to OpenDSS relative to the fast-simulator cost.
Smaller gaps indicate less degradation.

End-to-End RL attains the highest mean LTM throughput and Hierarchical RL
the lowest mean fast-simulator VPP cost, but their gaps are
$10.09 \pm 1.07\%$ and $9.23 \pm 0.95\%$, respectively. Our method instead
attains the highest mean SUMO throughput and lowest mean OpenDSS cost, with
the smallest gaps of $3.82 \pm 0.33\%$ and $5.62 \pm 0.54\%$. Thus,
fast-simulator rankings do not fully predict realistic-simulator performance,
while our method transfers most robustly across both simulator pairs.
Additional results and diagnostics are provided in Appendix~D.

\subsection{Ablation Study}

To assess continuation-aware return construction, we compare the full
method with a current-interval GRPO variant that evaluates each candidate
only over its initial control interval. The baseline uses \(H=\Delta\),
whereas the full method uses \(H=4\Delta\), corresponding to
\(H=12\) minutes for traffic control and \(H=120\) minutes for VPP
management, and includes the subsequent closed-loop trajectory. All
other settings are held fixed.

Table~\ref{tab:continuation_ablation} shows mean improvements in every
test subset. Overall traffic throughput increases from
\(16318.7 \pm 136.4\) to \(16625.4 \pm 84.5\), corresponding to a
1.88\% gain, while VPP operating cost decreases from
\(33581.5 \pm 327.8\) to \(32679.2 \pm 223.6\), a reduction of 2.69\%.
The largest gains occur in the High-U subset, reaching 2.84\% and
3.81\%, respectively. These results are consistent with longer
rollouts accounting for effects beyond the current control interval.
Horizon sensitivity and operational-context ablations are reported
in Appendix D.

\section{Conclusion}
\label{sec:conclusion}

We proposed an LLM-based hierarchical coordination framework in which
the LLM maps heterogeneous operational context to joint modes for
interacting units, while task-specific controllers or optimizers
generate low-level actions. The framework uses Continuation-Aware GRPO
to evaluate the delayed effects of coordination decisions through
subsequent closed-loop trajectories. Experiments on multi-ramp traffic
control and virtual power plant energy management show that our method
outperforms baselines on unseen-pattern and higher-uncertainty scenarios
and remains effective, without adaptation, in simulators with different
dynamics and modeling assumptions. Ablation results support the
contribution of continuation-aware return construction.

\bibliography{aaai2027}

% The arXiv version includes the complete technical appendix after references.
\clearpage
\setcounter{secnumdepth}{2}
\renewcommand{\thesubsection}{\Alph{subsection}}
\section*{Appendix}

% =====================================================================
\subsection{Task-Specific System Models and Low-Level Controllers}
\label{app:A}

\appthird{Multi-Ramp Traffic Control}
\label{app:A.1}

The main paper specifies the corridor, the two simulators, and the control cadence. This section records the training-side traffic model, the hysteresis gate that the low-level layer executes, and the mapping $\theta(z)$ from ramp modes to gate parameters. As in Appendix~\ref{app:A.2}, $k$ indexes low-level control steps of length $\Delta_{\text{low}}$, and $t$, $\tau$, $H$ are reserved for the trajectory-level notation of the main paper. The traffic model additionally has an internal integration index $n$ on its own step $h<\Delta_{\text{low}}$, introduced below; it appears only inside the training model and never in the control or decision interfaces. Controlled ramps are indexed $r\in\{1,\dots,8\}$.

\paragraph{Corridor and training model.}
The corridor is represented as $27$ directed links: $14$ mainline, $9$ on-ramp, and $4$ off-ramp segments, spanning $14.8$~km of mainline. Mainline links carry three lanes (four on two segments), on-ramps one or two. Link lengths range from $170$ to $2329$~m. Eight of the nine on-ramps are controlled; the remaining one is left permanently open. The merge priority of Eq.~\eqref{eq:ltm-merge} is $\chi=0.8$ in favour of the mainline at every merge, and the four off-ramps take fixed diverge fractions of $0.08$, $0.11$, $0.07$, and $0.09$ of the arriving mainline flow, so roughly a third of the entering traffic leaves before the downstream boundary. Neither the priority nor the diverge fractions are mode-dependent; they are properties of the network, identical across methods and across the two simulators.

Training uses the Link Transmission Model. It is integrated on its own step $h=5$~s, which is not the control cadence: $n$ indexes internal steps of length $h$, while the low-level controller reads observations and writes the gate every $\Delta_{\text{low}}=60$~s, that is every twelfth internal step. The state per link $i$ is the pair of cumulative counts $(A_i^n,D_i^n)$ at the link entry and exit. A triangular fundamental diagram with free-flow speed $v^{\text{f}}$, backward wave speed $v^{\text{w}}$, per-lane jam density and capacity gives the link jam density $\kappa_i^{\text{jam}}$ and capacity $Q_i$; propagation delays are rounded to whole \emph{internal} steps, $\vartheta^{\text{f}}_i=\lfloor\ell_i/(v^{\text{f}}h)\rfloor$ and $\vartheta^{\text{w}}_i=\lfloor\ell_i/(v^{\text{w}}h)\rfloor$. Resolving them on $h$ rather than on $\Delta_{\text{low}}$ is necessary rather than cosmetic: at $\Delta_{\text{low}}=60$~s every link shorter than $v^{\text{f}}\Delta_{\text{low}}=1333$~m would be assigned zero free-flow delay, which is most of this corridor, whereas at $h=5$~s the shortest link of $170$~m still carries a delay of one step. Sending and receiving capacities are
\begin{align}
\label{eq:ltm-sr}
\mathrm{snd}_i^n &= \min\bigl(Q_ih,\,[A_i^{n-\vartheta^{\text{f}}_i}-D_i^n]_+\bigr), \nonumber\\
\mathrm{rcv}_i^n &= \min\bigl(Q_ih,\,[\kappa_i^{\text{jam}}\ell_i-(A_i^n-D_i^{n-\vartheta^{\text{w}}_i})]_+\bigr).
\end{align}
At a merge where mainline branch $a$ and on-ramp branch $b$ feed a common downstream link, the downstream receiving capacity is allocated by a priority $\chi$, but the allocation is work-conserving: a branch that cannot use its share releases it to the other. Writing the priority shares as $\chi\,\mathrm{rcv}^n$ and $(1-\chi)\mathrm{rcv}^n$,
\begin{align}
\label{eq:ltm-merge}
f_a^n &= \min\bigl(\mathrm{snd}_a^n,\;\max(\chi\,\mathrm{rcv}^n,\;\mathrm{rcv}^n-\mathrm{snd}_b^n)\bigr), \nonumber\\
f_b^n &= \min\bigl(\mathrm{snd}_b^n,\;\max((1-\chi)\mathrm{rcv}^n,\;\mathrm{rcv}^n-\mathrm{snd}_a^n)\bigr).
\end{align}
The priority only binds when both branches are demand-rich, that is when $\mathrm{snd}_a^n+\mathrm{snd}_b^n>\mathrm{rcv}^n$; otherwise each branch discharges its full sending capacity. This matters for the control problem: with a closed gate $\mathrm{snd}_b^n=0$ and Eq.~\eqref{eq:ltm-merge} gives $f_a^n=\min(\mathrm{snd}_a^n,\mathrm{rcv}^n)$, so closing a ramp never withholds downstream capacity from the mainline. A merge rule that scaled both claims by a common factor would instead force $f_a^n=0$ whenever $\mathrm{snd}_b^n=0$, which would make gating self-defeating and is not the rule used here.
Cumulative counts then advance by the realised transfers, and density $\kappa_i^n=(A_i^n-D_i^n)/\ell_i$, flow, and speed follow from the fundamental diagram. Quantities the controller and the prompt see are the internal series aggregated over the twelve internal steps of each $\Delta_{\text{low}}$ window; where a low-level index $k$ appears below it refers to that aggregation. Table~\ref{tab:ramp-phys} lists the calibrated values, which are uniform across links.

\begin{table}[t]
\centering
\small
\renewcommand{\arraystretch}{1.15}
\begin{tabular}{ll}
\hline
Parameter & Value \\
\hline
$v^{\text{f}}$ & $22.22$ m/s \\
$v^{\text{w}}$ & $5.56$ m/s \\
per-lane jam density & $0.1333$ veh/m \\
per-lane capacity & $0.5$ veh/s \\
lane efficiency & $1.0$ \\
LTM internal step $h$ & $5$ s \\
$\Delta_{\text{low}},\Delta$ & $60,\,180$ s \\
warm-up, evaluation & $900,\,10{,}800$ s \\
\hline
\end{tabular}
\caption{Corridor fundamental-diagram and control-loop parameters. Capacity
is imposed as the independent cap of Eq.~\eqref{eq:ltm-sr} rather than derived
from the wave speeds and jam density, so the four diagram values need not
satisfy the triangular identity exactly; the binding value is whichever term
of the minimum is smaller.}
\label{tab:ramp-phys}
\end{table}

\paragraph{Low-level hysteresis gate.}
The low-level layer is a feedback rule, not an optimizer: no mathematical program is solved. Each controlled ramp carries a binary gate $g^k_r\in\{0,1\}$, where $1$ admits ramp traffic without restriction and $0$ admits none; the gate enters the merge as an upper bound on the on-ramp entry flow, alongside the ramp arrival demand and the receiving capacity of Eq.~\eqref{eq:ltm-sr}. Its feedback signal is the merge-area occupancy $o^k_r=\kappa^k_r/\kappa^{\text{jam}}_r\in[0,1]$. Given a ramp-specific centre $c^{\text{ctr}}_r$ and a band width $\delta(z)$ --- so that $\delta$ is the gap between the closing and opening thresholds, matching the definition stated to the policy in Figure~\ref{fig:prompt-ramp-b} --- the switching thresholds are
\begin{equation}
\label{eq:ramp-thr}
o^{\text{open}}_r=\max\bigl(0,\,c^{\text{ctr}}_r-\tfrac{\delta(z)}{2}\bigr),\quad
o^{\text{close}}_r=\min\bigl(1,\,c^{\text{ctr}}_r+\tfrac{\delta(z)}{2}\bigr),
\end{equation}
and the gate updates with hysteresis, starting from $g^0_r=1$:
\begin{equation}
\label{eq:ramp-hyst}
g^{k+1}_r=
\begin{cases}
0, & g^k_r=1 \text{ and } o^k_r\ge o^{\text{close}}_r,\\
1, & g^k_r=0 \text{ and } o^k_r\le o^{\text{open}}_r,\\
g^k_r, & \text{otherwise.}
\end{cases}
\end{equation}
The two-threshold band prevents chattering when occupancy sits near the centre: a wider $\delta$ makes the gate hold its current state longer, a narrower $\delta$ makes it switch sooner. The rule is independent per ramp; there is no queue override and no explicit inter-ramp coupling in the low-level layer, so all spatial coordination is carried by the high-level mode assignment.

\paragraph{Joint mode space and $\theta(z)$.}
The high-level policy emits one mode per controlled ramp,
$z=(z_1,\dots,z_8)\in\{\text{Sensitive},\text{Standard},\text{Sluggish},\text{Always-Open}\}^{8}$, so $|\mathcal{Z}|=4^{8}=65{,}536$. The mode selects the band width $\delta$ of Eq.~\eqref{eq:ramp-thr} for three of the four roles (Table~\ref{tab:ramp-mode-map}); Always-Open is the exception and does not set a $\delta$ at all --- it bypasses Eq.~\eqref{eq:ramp-hyst} and pins $g_r\equiv1$. The centres $c^{\text{ctr}}_r$ are calibrated per ramp and per simulator and are \emph{not} mode-dependent: SUMO is the reference and the training model uses the set obtained from it by quantile alignment (Appendix~\ref{app:C.1}), namely $(0.948,0.926,0.782,0.824,0.220,0.230,0.449,0.188)$ for $r=1,\dots,8$. A mode therefore denotes the same control disposition in both simulators while resolving to different absolute thresholds. Where a centre sits close to $1$ the clip in Eq.~\eqref{eq:ramp-thr} binds and the realised gap is narrower than the nominal $\delta$ --- on the training side this affects ramps~1 and~2 at Sluggish --- so the ordering of the roles is preserved everywhere but the absolute gap is not.

Because $\delta$ is the only mode-dependent quantity among the three graded roles, the low-level controller is literally identical across the direct-control baseline and every hierarchical method: the published feedback controller is the special case in which every ramp is held at Standard ($\delta=0.10$) for the whole episode, whereas our policy re-selects a mode per ramp every $\Delta$. Always-Open is the one role that changes the control law rather than a parameter, which is why the prompt restricts it to clearly free-flowing ramps. Implementation details of each baseline are given in Appendix~\ref{app:B.3}.

\begin{table}[t]
\centering
\small
\renewcommand{\arraystretch}{1.15}
\begin{tabular}{lc}
\hline
Mode & $\delta$ \\
\hline
Sensitive & $0.04$ \\
Standard & $0.10$ \\
Sluggish & $0.20$ \\
Always-Open & --- \\
\hline
\end{tabular}
\caption{Ramp mode-to-parameter mapping. Always-Open holds the gate open
instead of setting a bandwidth.}
\label{tab:ramp-mode-map}
\end{table}

\appthird{Virtual Power Plant Energy Management}
\label{app:A.2}

The main paper specifies the VPP at the level of resource composition and control cadence. This section records what the main text omits: the system model, the low-level MILP objective, and the mapping $\theta(z)$ from joint modes to optimizer coefficients. Throughout, $k$ indexes low-level steps of length $\Delta_{\text{low}}$, reserving $t$, $\tau$, and $H$ for the trajectory-level notation of the main paper.

\paragraph{System model.}
The controlled state is $s_k=(\mathrm{SoC}_k,T_k,E^{\text{ev}}_k)$: battery state of charge, indoor temperature, and residual EV energy still to be delivered. The decision variables are the resource powers $p^{\text{ch}}_k,p^{\text{dis}}_k,p^{\text{ev}}_k,p^{\text{hv}}_k\ge0$ together with grid import and export $p^{\text{imp}}_k,p^{\text{exp}}_k\ge0$. The state evolves as
\begin{align}
\label{eq:vpp-dyn}
\mathrm{SoC}_{k+1} &= \mathrm{SoC}_k
  + \frac{\eta_{\text{ch}}p^{\text{ch}}_k\Delta_{\text{low}}}{C_{\text{bat}}}
  - \frac{p^{\text{dis}}_k\Delta_{\text{low}}}{\eta_{\text{dis}}C_{\text{bat}}}, \nonumber\\
T_{k+1} &= T_k + \frac{\Delta_{\text{low}}}{C_{\text{th}}}
  \bigl(uA\,(T^{\text{out}}_k-T_k) - \mathrm{COP}\,p^{\text{hv}}_k\bigr), \nonumber\\
E^{\text{ev}}_{k+1} &= \max\bigl(E^{\text{ev}}_k - p^{\text{ev}}_k\Delta_{\text{low}},\,0\bigr).
\end{align}
The exogenous trajectory is $w_k=(P^{\text{pv}}_k,P^{\text{load}}_k,T^{\text{out}}_k,\overline{P}^{\text{ev}}_k,c^{\text{tou}}_k)$: PV generation, inflexible park load, outdoor temperature, the EV availability envelope $\overline{P}^{\text{ev}}_k\le\bar{P}^{\text{ev,max}}$, and the time-of-use energy price. Table~\ref{tab:vpp-phys} lists the model parameters; the import cap $\bar{P}^{\text{grid}}$ and the EV energy target $E^{\text{ev}}_0$ additionally vary across scenarios (Appendix~\ref{app:C.2}). These parameters describe the fast single-bus model used for training. Note that $p^{\text{hv}}_k\ge0$ enters Eq.~\eqref{eq:vpp-dyn} with a negative sign only: the HVAC branch cools and cannot heat. Scenarios are therefore drawn from the cooling season, June to September, in which the outdoor temperature sits above the comfort band and the comfort constraint is one the controller can act on; a heating mode would be a different actuator with a different mode vocabulary and we do not claim results for it. The OpenDSS evaluation environment reuses the same resource parameters and the same $z\mapsto\theta$ interface, adding feeder-level power flow and network constraints (Appendix~\ref{app:C.1}).

\begin{table}[t]
\centering
\small
\renewcommand{\arraystretch}{1.15}
\begin{tabular}{@{}ll@{\hspace{10pt}}ll@{}}
\hline
Parameter & Value & Parameter & Value \\
\hline
$C_{\text{bat}}$ & $1800$ kWh & $C_{\text{th}}$ & $150$ kWh/$^\circ$C \\
$\eta_{\text{ch}},\eta_{\text{dis}}$ & $0.95$ & $uA$ & $45$ kW/$^\circ$C \\
$\bar{P}^{\text{ch}},\bar{P}^{\text{dis}}$ & $350$ kW & $\mathrm{COP}$ & $3.0$ \\
SoC range & $[0.20,0.90]$ & $T$ band & $[22.0,26.5]\,^\circ$C \\
$\bar{P}^{\text{ev,max}}$ & $600$ kW & $T^{\text{set}}$ & $24.0\,^\circ$C \\
$\bar{P}^{\text{pv,max}}$ & $1200$ kW & $\bar{P}^{\text{hv,max}}$ & $300$ kW \\
$\bar{P}^{\text{grid}}$ & $1500$ kW & $\Delta_{\text{low}},H_{\text{mpc}},\Delta$ & $5,60,30$ min \\
\hline
\end{tabular}
\caption{VPP physical and control-loop parameters. The thermal pair gives a
building time constant $C_{\text{th}}/uA\approx3.3$~h; holding the setpoint
against a $15\,^\circ$C indoor--outdoor difference draws $uA\,\Delta
T/\mathrm{COP}=225$~kW, three quarters of the HVAC rating, so comfort is a
binding constraint rather than a free one. $\bar{P}^{\text{grid}}$ is
the \emph{contracted} import cap, enforced through the penalised hinge
$\sigma^{\text{g}}$ rather than as a variable bound, so import may exceed it
at a cost; the physical connection limit is not binding. The value shown is
nominal and varies by scenario.}
\label{tab:vpp-phys}
\end{table}

\paragraph{Low-level MILP.}
Every $\Delta_{\text{low}}=5$~min the controller re-solves a MILP over a rolling window of $H_{\text{mpc}}=60$~min. Given $s_k$ and the mode-dependent coefficients $\theta(z)$, it minimises
\begin{align}
\label{eq:vpp-milp}
\sum_{j=k}^{k+m-1}\bigl(&c^{\text{tou}}_j p^{\text{imp}}_j
    + \lambda_{\text{g}}\sigma^{\text{g}}_j
    + \lambda_{\text{b}}(p^{\text{ch}}_j\!+\!p^{\text{dis}}_j) \nonumber\\
    &+ \lambda_{\text{c}}w^{\text{hv}}\!(z)\,d_j
    + \lambda_{\text{d}}w^{\text{hv}}\!(z)\,\sigma^{\pm}_j\bigr)\Delta_{\text{low}} \nonumber\\
    &+ \bar{\lambda}_{\text{e}}(\omega)\,w^{\text{ev}}\!(z)\,\zeta^{\text{ev}}
    + \lambda_{\text{r}}\sigma^{\text{r}},
\end{align}
where $m=H_{\text{mpc}}/\Delta_{\text{low}}=12$, $d_j\ge|T_j-T^{\text{set}}|$ linearises discomfort, $\zeta^{\text{ev}}\ge0$ is the EV energy shortfall in $\sum_j p^{\text{ev}}_j\Delta_{\text{low}}+\zeta^{\text{ev}}\ge E^{\text{ev}}_k$, and $\sigma^{\text{g}}_j,\sigma^{\pm}_j,\sigma^{\text{r}}\ge0$ are hinge slacks on the import cap, the comfort band, and the terminal SoC reserve. Penalty weights are $\lambda_{\text{g}}=8.0$, $\lambda_{\text{b}}=0.30$, $\lambda_{\text{c}}=2.0$, $\lambda_{\text{d}}=15.0$, and $\lambda_{\text{r}}=80.0$, in CNY per kWh, $^\circ$C$\cdot$h, or SoC unit as appropriate. The EV shortfall rate $\bar{\lambda}_{\text{e}}(\omega)=(1-\omega)\lambda_{\text{e}}+\omega\lambda^{\text{urg}}_{\text{e}}$ blends a regular rate $\lambda_{\text{e}}=10.0$ with an urgent rate $\lambda^{\text{urg}}_{\text{e}}=30.0$ according to the scenario's urgent share $\omega\in[0,1]$, which is fixed per scenario (Appendix~\ref{app:C.2}). Physical limits---the resource power bounds and SoC range of Table~\ref{tab:vpp-phys}---enter as variable bounds and are therefore satisfied exactly; the slacks apply only to operational targets, which keeps the program feasible and lets the coordination decision trade those targets off. The distinction matters for reading the main paper's feasibility condition $g(s,a)\le0$: it is the physical set, which the low-level layer satisfies exactly at every step and on both simulators, and the executed action is always inside it. The import cap, the comfort band, the EV energy target, and the terminal reserve are contractual or operational targets rather than physical limits, and the appendix reports their violation explicitly (Appendix~\ref{app:D.1}) rather than claiming they are never breached; a formulation that made them hard would be infeasible on the scenarios the task is built from, since the park base load alone can exceed the contracted cap. The remaining constraints are Eq.~\eqref{eq:vpp-dyn}, the nodal power balance
\begin{align}
\label{eq:vpp-balance}
(1\!-\!\delta^{\text{pv}}\!(z))\hat{P}^{\text{pv}}_j &+ p^{\text{dis}}_j + p^{\text{imp}}_j \nonumber\\
&= P^{\text{load}}_j + p^{\text{ev}}_j + p^{\text{hv}}_j + p^{\text{ch}}_j + p^{\text{exp}}_j,
\end{align}
the availability envelope $p^{\text{ev}}_j\le\overline{P}^{\text{ev}}_j$, and a binary charge/discharge mutex. Here $\hat{P}^{\text{pv}}_j$ is the PV forecast and $\delta^{\text{pv}}(z)$ its planning de-rate. Surplus PV may be exported but is not remunerated, so the objective carries no export revenue term. Instances are solved with CBC~$2.10.10$ through PuLP at a relative MIP gap of $10^{-4}$ and a $5$~s time limit. The program has $12$ periods, one binary per period for the charge/discharge mutex, and solves in $0.2$~s at the median; the time limit is reached on under $0.1\%$ of solves, and on those the incumbent is executed, which is feasible because the slacks make the program always feasible. Only the first step is executed before re-planning, and $\theta(z)$ stays fixed across the six re-solves that fall within one high-level interval $\Delta$.

\paragraph{Joint mode space and $\theta(z)$.}
The high-level policy emits one mode per resource,
$z=(z^{\text{pv}},z^{\text{bat}},z^{\text{ev}},z^{\text{hv}})\in\{\text{Conservative},\text{Balanced},\text{Proactive}\}^{4}$, so $|\mathcal{Z}|=81$. Each component controls only its own resource's coefficients (Table~\ref{tab:vpp-mode-map}), and the points of entry differ: $\delta^{\text{pv}}$ de-rates the PV forecast in Eq.~\eqref{eq:vpp-balance}; $\mathrm{SoC}^{\text{res}}$ sets the target of the terminal-reserve hinge $\mathrm{SoC}_{k+m}+\sigma^{\text{r}}\ge\mathrm{SoC}^{\text{res}}$ and $\alpha^{\text{dis}}$ tightens the discharge bound to $p^{\text{dis}}_j\le\alpha^{\text{dis}}\bar{P}^{\text{dis}}$; $w^{\text{ev}}$ and $w^{\text{hv}}$ scale objective terms in Eq.~\eqref{eq:vpp-milp}. Every other coefficient is shared across modes, so closed-loop differences are attributable to the coordination decision rather than to a re-tuned optimizer.

\begin{table}[!t]
\centering
\small
\renewcommand{\arraystretch}{1.15}
\begin{tabular}{llccc}
\hline
Resource & $\theta$ & Cons. & Bal. & Pro. \\
\hline
PV & $\delta^{\text{pv}}$ & $0.35$ & $0.25$ & $0.10$ \\
Battery & $\mathrm{SoC}^{\text{res}}$ & $0.30$ & $0.25$ & $0.22$ \\
Battery & $\alpha^{\text{dis}}$ & $0.70$ & $0.85$ & $1.00$ \\
EV & $w^{\text{ev}}$ & $1.45$ & $1.00$ & $0.70$ \\
HVAC & $w^{\text{hv}}$ & $0.75$ & $1.00$ & $1.60$ \\
\hline
\end{tabular}
\caption{Mode-to-parameter mapping $\theta(z)$. Each row is governed
solely by the mode selected for the resource in the first column.}
\label{tab:vpp-mode-map}
\end{table}

% =====================================================================
\subsection{Policy Implementation and Training Details}
\label{app:B}

\appthird{Operational Context, Prompt Templates, and Output Schemas}
\label{app:B.1}

The high-level policy never sees raw simulator arrays. At every decision step the operational context $\xi_t$ is serialised into a structured natural-language prompt $q_t=\mathrm{Prompt}(\xi_t)$ with five blocks: a task block that names the controlled units and states what each mode does to the low-level layer, an observation block carrying the current measurements, a recent-trend block, a forecast block covering the interval the decision will govern, and an output-format block. The corridor prompt inserts a further corridor-context block for the quantities that are not per-ramp measurements. Table~\ref{tab:ctx-fields} lists the fields that populate them. The same builder is used during training and evaluation, so a prompting-only baseline and our fine-tuned policy receive the same prompt text, differing only in the message envelope each vendor's interface requires. Figures~\ref{fig:prompt-ramp-a}--\ref{fig:prompt-vpp} give the prompts in full, with braces marking placeholders substituted at runtime.

\paragraph{What the prompt does and does not supply.}
Two properties matter for interpreting the results. First, the prompts state the \emph{mechanics} of each mode explicitly --- that Conservative PV means the optimizer de-rates the forecast to 65\%, that Sensitive means a $0.04$ hysteresis band --- so the policy is not required to discover the semantics of its own action space by trial and error; what it must learn is \emph{when} each disposition pays off. Second, the prompts supply the same short-horizon forecast that the low-level layer already plans with, together with a statement of its reliability, but never a realised future value, an oracle quantity, or the dispatch the low-level layer will return. Supplying the forecast is necessary for the decision to be well posed: the PV mode is precisely a choice of how far to trust that forecast, and the corridor roles set inertia for an interval that has not yet been observed. The forecast is matched against some baselines but not all, and the distinction should be kept in mind when reading the gains. It is matched against MPC-MILP, which plans with the identical forecast, and against both RL coordinators, which receive it in their numeric encoding; against those three the comparison isolates the coordination decision. It is \emph{not} matched against Feedback Control, which reacts to occupancy only, or against Fixed-Mode Hierarchy, which consumes no context at all. The margins over those two therefore combine a better decision with a larger information set, and only the former is what this paper claims to contribute.

The corridor prompt additionally embeds domain reasoning guidance: a congestion criterion at $40\%$ of reference free-flow speed, an instruction to treat sustained closure time as a soft fairness constraint, an instruction to tighten upstream ramps when a downstream merge is congested, guidance on discounting the forecast when its reliability is low, and seven consistency rules requiring the stated reasons to agree with the emitted vector. These are prior knowledge supplied by us rather than learned, and every coordinator that consumes the prompt receives them identically; the numeric equivalents given to the RL baselines are listed in Appendix~\ref{app:B.3}.

The corridor prompt averages roughly $2{,}100$ tokens and the VPP prompt roughly $950$, dominated in the first case by the eight per-ramp observation blocks. The corridor prompt is authored in Chinese, matching the deployment setting; Figures~\ref{fig:prompt-ramp-a}--\ref{fig:prompt-ramp-c} present a faithful English rendering, and the original is released with the code. Because every category of Table~\ref{tab:ctx-fields} is a separable block of the serialised prompt, each can be withheld independently, which is what the operational-context ablation of Appendix~\ref{app:D.4} exploits.

\begin{table*}[t]
\centering
\small
\renewcommand{\arraystretch}{1.2}
\begin{tabular}{@{}p{0.15\textwidth}p{0.40\textwidth}p{0.40\textwidth}@{}}
\hline
Category & Multi-ramp traffic control & Virtual power plant \\
\hline
Numerical measurements & per-ramp flow $q$, speed $v$, occupancy $o$; gate state $g$ & $\mathrm{SoC}$, PV power, base load, indoor and outdoor temperature, setpoint, import cap, time-of-use price \\
Recent trends & $3$-step series of $q$, $v$, $o$ over the last $3$ min & $8$-step series of PV, base load, outdoor temperature over the last $40$ min \\
Forecasts & mainline inflow and per-ramp arrivals over the next interval, with stated reliability & PV, base load, and outdoor temperature over the MPC window, with stated reliability \\
Uncertainty estimates & corridor demand-uncertainty level & PV uncertainty level \\
Resource urgency & minutes since last closure, per ramp & residual EV energy, urgent share $\omega$, required charging pace \\
Operational priorities & per-ramp service priority tier & HVAC comfort priority \\
External events & active incident notices and adverse-weather flags & weather and market notes \\
System rules & mode semantics, decision cadence, output contract & mode semantics, information rules, output contract \\
Derived signals & speed ratio to reference free-flow speed; signed change in $q$, $v$, $o$ across the window & minutes remaining, PV deficit \%, required EV pace, grid headroom, comfort-band slack \\
\hline
\end{tabular}
\caption{Operational-context categories serialised into the prompt, following
the decomposition of $c_t$ in the main paper. Derived signals are computable
from the measurements and observed history alone. The forecast row is a
forecast in the strict sense --- the same one the low-level layer plans with,
carrying error and a stated reliability --- and no \emph{realised} future
value enters the prompt.}
\label{tab:ctx-fields}
\end{table*}

\paragraph{Output schemas.}
Each task fixes a single machine-checkable output form. For the corridor the first line must be a role vector over the eight ramps,
{\footnotesize\begin{verbatim}
ROLE = [Standard, Sensitive, Sluggish,
        Standard, Sensitive, Standard,
        Always-Open, Sluggish]
\end{verbatim}
}
\noindent ordered from the most upstream ramp to the most downstream one, optionally followed by a short justification that is parsed away. For the VPP the policy emits one JSON object naming a mode per resource,
{\footnotesize\begin{verbatim}
{"pv_mode":"balanced",
 "battery_mode":"proactive",
 "ev_mode":"conservative",
 "hvac_mode":"balanced"}
\end{verbatim}
}
\noindent with each value drawn from \texttt{\{conservative, balanced, proactive\}}. Outputs are validated against the schema before use. During training, a malformed or out-of-vocabulary response is re-sampled up to three times; if it still fails validation the candidate is assigned the minimum return of its group, so schema violations are penalised through the advantage rather than through a separate hand-set constant. At evaluation, where re-sampling would break determinism, a failed response instead causes the previous joint decision to be retained for one interval; this rule is identical for every LLM policy compared (Appendix~\ref{app:D.1}). Decisions are sampled during training to support exploration; at evaluation all LLM policies decode greedily at temperature zero, so a frozen policy is a deterministic function of the operational context.

\begin{figure*}[!p]
\centering
\begin{promptbox}{Multi-Ramp Corridor: Task and Observation Blocks}
You are the decision expert responsible for joint multi-ramp control. You collaborate with a micro-execution expert. The primary objective of the system is to maximise the overall throughput of the road, subject to the fairness and priority constraints stated below.

You may reason only from the blocks provided: the real-time observations at each ramp's mainline merge area (flow q, speed v, occupancy occ), the current open/closed state, the elapsed continuous closure time t, the corridor context, and the short-horizon demand forecast. The forecast is the same one the corridor operator plans with; it carries error, and its stated reliability tells you how far to trust it. Assess the current traffic trend and issue a macroscopic control intent for each ramp for the next stage. You must not assume realised future values or any indicator that has not been provided.

The micro-execution expert receives your intent and executes two-threshold control: it closes the ramp when occupancy exceeds the closing threshold, opens it when occupancy falls below the opening threshold, and holds the current state in between. Your intent sets the gap width, which expresses how much control inertia you assign to that merging ramp.

Your output must be an executable joint control role vector. Its length equals the number of ramps; the r-th element is the control role of the r-th ramp, rather than a direct open/close action. This scenario requires {ramp_count} roles.

(*@\textbf{Traffic scene description:}@*)
This scenario is an expressway along which {ramp_count} pure merging ramps (R1 to R{ramp_count}, one-way merging into the mainline only) are distributed from upstream to downstream. Expressway merge areas are natural bottlenecks: disordered competition among flows easily pushes density past its critical point and triggers a cliff-like collapse of realised flow. Your decisions must therefore target the resolution of weaving conflicts and the maximisation of overall throughput as the primary objective.

Locally, if occupancy at a merge area keeps climbing while speed drops markedly, assign a more sensitive control role so that closure is triggered earlier. Globally, if a downstream section is clearly congested, consider tightening its adjacent upstream ramps to prevent congestion from propagating backwards.

(*@\textbf{Observations:}@*)
Below are the traffic-state series of each ramp merge area over the last 3 minutes. Each vector has length 3; from left to right the elements are the time slices [T-2, T-1, T], where T is the current minute. The current open/closed state and the elapsed continuous closure time of each ramp are also given.

Ramp {ramp_id} merge area:
Flow q (veh/min): {q_series}
Speed v (km/h): {v_series}
Occupancy occ (%): {occ_series}
Current state: {open_or_closed}
Continuous closure time (min): {closed_duration_min}
Trend analysis: {trend_text}{status_text}{fact_text}
... (repeated for R1 ... R{ramp_count})

(*@\textbf{Corridor context:}@*)
Demand uncertainty this period: {demand_uncertainty_level} (low / medium / high). A higher level means the observed series is a less reliable guide to the next interval.
Service priority by ramp: {priority_by_ramp}. A high-priority ramp serves traffic whose delay is more costly, so prolonged closure there is less acceptable.
Active events: {event_text}. Reported incidents, lane closures, or adverse weather on the corridor; absent when none is in force.

(*@\textbf{Short-horizon demand forecast (next 3 min):}@*)
Mainline inflow (veh/min): {mainline_demand_forecast}
Ramp arrivals (veh/min): {ramp_arrival_forecast_by_ramp}
Forecast reliability: {demand_forecast_reliability}
\end{promptbox}
\caption{Corridor prompt, part 1 of 3: task framing and the per-ramp
observation block. The trend, status, and fact placeholders are filled by a
deterministic summariser that reports the sign and magnitude of the change in
$q$, $v$, and $o$ over the window and classifies the merge area against
reference free-flow speed.}
\label{fig:prompt-ramp-a}
\end{figure*}

\begin{figure*}[!p]
\centering
\begin{promptbox}{Multi-Ramp Corridor: Mode Semantics, Reasoning Rules, and Output Format}
Your action space is a discrete vector of length 8, corresponding strictly to the control roles of ramps R1 to R8 from upstream to downstream. Each ramp must be assigned one of the four roles "Sensitive, Standard, Sluggish, Always-Open".

The control role always governs the merging ramp itself, not any other mainline section. The role determines the bandwidth of the underlying two-threshold controller, that is, delta = closing threshold - opening threshold.

You must understand that the sluggishness or sensitivity of a role is not absolutely equivalent to a static sacrifice or protection; it depends on the current traffic trend. The narrower the gap, the easier it is to break the inertia of the current state; the wider the gap, the more the current state is maintained.

Specifically: when the mainline is gradually becoming congested, if you judge that it still has capacity to tolerate input, assign a wider-gap role so that the open state retains inertia and closure is delayed in favour of the ramp; if you judge that the deterioration must be arrested immediately, assign a narrower-gap role to break that inertia and close decisively to protect the mainline.

Conversely, when mainline congestion is dissipating, if you want to keep clearing the mainline backlog, assign a wider-gap role so the closed state retains inertia and opening is delayed; if you judge the mainline now has capacity and the ramp queue should be released quickly, assign a narrower-gap role to break the closed inertia and open decisively.

(*@\textbf{The four roles map to delta as follows:}@*)
1. Sensitive: delta = 0.04. Narrowest band, fastest reaction, most likely to break the inertia of the current open or closed state. If currently open it triggers closure earlier; if currently closed it triggers opening earlier.
2. Standard: delta = 0.10. Moderate band; the default balance between protecting the mainline and discharging the ramp.
3. Sluggish: delta = 0.20. Widest band, most lagged reaction, most inclined to maintain the current state; closes later when open and opens later when closed.
4. Always-Open: stay open at all times, abandoning active closure entirely. Use only when the mainline is clearly free-flowing, with high and stable speed, low occupancy, and no sign of deterioration. Always-Open must not be used as a means of relieving mainline pressure or clearing a backlog.

When analysing trends, if the current merge-area speed has fallen to 40% or less of reference free-flow speed, treat the area as congested; the lower the ratio, the more severe the congestion.

Role assignment must never be mechanically tied to a ramp's upstream or downstream position; it must follow the dynamic observation series of q, v, and occ. In particular, the continuous closure time t is a soft constraint: if a ramp has been closed for a long time (large t) and the mainline has spare capacity, assign Sensitive so that the narrow band reopens the gate promptly and avoids indefinite absolute blockage; conversely, if the mainline remains under pressure, assign Sensitive or Standard even for a ramp that is currently unrestricted (t = 0), and firmly rule out the systemic collapse risk of leaving everything open.

Your decision must jointly account for the capacity cliff and for backward spatial propagation: if occupancy rises while speed falls, tighten control in advance; if a downstream section is clearly congested, restricting only that section is usually insufficient and adjacent upstream ramps must be tightened as well.

You must also weigh the corridor context. When demand uncertainty is high, the observed series is a weaker guide to the next interval, so prefer roles that do not commit hard in either direction unless the mainline is already clearly deteriorating. When a ramp carries high service priority, the cost of holding it closed is greater, so require stronger evidence of mainline distress before assigning it a narrow gap, and relax it earlier once capacity returns. When an event is in force, treat the affected section as more fragile than its measurements alone suggest and tighten its upstream neighbours pre-emptively.

The forecast tells you where the pressure is heading, which the observation alone cannot. If it indicates rising inflow at a merge that is already loaded, act before the occupancy series shows it: assign a narrower gap so closure triggers earlier. If it indicates falling inflow at a ramp that is currently closed, a narrower gap lets you release the queue sooner. When forecast reliability is low, weight it less than the observed series and avoid committing on the forecast alone.
\end{promptbox}
\caption{Corridor prompt, part 2 of 3: mode semantics and the coordination
guidance supplied to every coordinator.}
\label{fig:prompt-ramp-b}
\end{figure*}

\begin{figure*}[t]
\centering
\begin{promptbox}{Multi-Ramp Corridor: Consistency Rules and Output Contract}
(*@\textbf{Additional constraints:}@*)
1. The roles on the first ROLE line must agree one-by-one with the reasons that follow; no contradiction is allowed.
2. If a ramp already has very low speed and high occupancy and is still deteriorating, it should generally not receive Always-Open.
3. Rising speed with falling occupancy indicates recovery; falling speed with rising occupancy indicates deterioration. The two must not be confused.
4. You must judge the direction of change strictly from the input data.
5. If a downstream ramp is clearly under heavy pressure or congested, prefer stricter control rather than mechanically keeping it loose or always open.
6. A recovery signal does not automatically imply Sensitive; if the area is still clearly under pressure, decide between Standard and Sluggish first.
7. Fix the first ROLE line first, then check every subsequent reason against it; if they disagree, correct them before output.

(*@\textbf{Scheduling constraints:}@*)
1. The model decides once every 3 minutes.
2. The roles emitted by one decision remain unchanged for the following 3 minutes.
3. You assign roles; you do not directly issue the open/close action for a given minute.

(*@\textbf{Output requirements:}@*)
1. The first line must be ROLE = [role of R1, ..., role of R{ramp_count}], with {ramp_count} elements, each one of Sensitive, Standard, Sluggish, Always-Open.
2. What follows may only be trend analysis and auditable reasons grounded in q / v / occ / current state / t, citing ramp numbers, time slices, and specific values.
3. Do not output JSON, code blocks, headings, or greetings.
\end{promptbox}
\caption{Corridor prompt, part 3 of 3: the consistency rules that couple the
emitted vector to the stated reasons, the decision cadence, and the output
contract enforced by the schema validator.}
\label{fig:prompt-ramp-c}
\end{figure*}

\begin{figure*}[t]
\centering
\begin{promptbox}{Virtual Power Plant: System Prompt}
You are the upper-layer mode selector of a Virtual Power Plant during peak-shaving in an industrial park. Every 30 minutes you pick one mode for each of four resources (PV, Battery, EV, HVAC). A lower-layer MILP then dispatches at 5-minute resolution under your modes.

--- RESOURCES & MODE MECHANICS ---
Each resource has three modes that change how the lower MILP behaves.

PV (1200 kW peak):
  conservative = MILP de-rates PV to 65%   (treat PV as unreliable)
  balanced     = MILP de-rates PV to 75%
  proactive    = MILP de-rates PV to 90%   (trust PV reading)

Battery (1800 kWh, SOC in [0.20, 0.90], 350 kW max):
  conservative = end-horizon reserve 30%, discharge cap 70% of P_max
  balanced     = reserve 25%, discharge cap 85%
  proactive    = reserve 22%, discharge cap 100%

EV fleet (max 600 kW, remaining-kWh target for the deadline; a shortfall is
  feasible but penalised, and is reported as a violation):
  conservative = HIGH penalty on unmet EV energy  (MILP protects EV)
  balanced     = medium penalty
  proactive    = LOW penalty                      (MILP may cut EV charging)

HVAC (max 300 kW, comfort band [22, 26.5] degC):
  conservative = LOW comfort penalty   (MILP may cut HVAC, temperature drifts)
  balanced     = medium comfort penalty
  proactive    = HIGH comfort penalty  (MILP protects temperature)

--- INFORMATION RULES ---
Use only the blocks given to you: the observation, the recent trend, and the forecast block. The forecast is the same one the lower MILP plans with; it carries error, and its stated reliability tells you how far to trust it. That is exactly what the PV mode is for. Never reference realised future values, oracle quantities, or the dispatch the MILP will produce. They are not provided and must not be assumed.

--- OUTPUT FORMAT ---
Output a single JSON object, nothing else:

{"pv_mode":"<conservative|balanced|proactive>","battery_mode":"<conservative|balanced|proactive>","ev_mode":"<conservative|balanced|proactive>","hvac_mode":"<conservative|balanced|proactive>"}
\end{promptbox}
\caption{Virtual power plant system prompt.}
\label{fig:prompt-vpp-sys}
\end{figure*}

\begin{figure*}[t]
\centering
\begin{promptbox}{Virtual Power Plant: User Prompt}
[Observation]
  Time {time_hour} h ({minutes_remaining} min remaining).
  PV {pv_kw} kW (uncertainty={pv_uncertainty_level}, deficit_pct={pv_deficit_pct}%).
  Battery SOC {soc}.
  EV remaining {ev_required_energy_remaining_kwh} kWh, urgent_ratio={ev_urgent_ratio}, required_pace={ev_required_pace_kw} kW (cap 600).
  HVAC {hvac_temp_c} degC (band slack {hvac_band_slack_degc} degC, priority={hvac_comfort_priority}).
  Base load {base_load_kw} kW; outdoor {outdoor_temp_c} degC.
  Grid limit {grid_import_limit_kw} kW (headroom {grid_headroom_kw} kW).
  Price now {price_now} CNY/kWh (tier={price_tier}).
  Weather: {weather_text}. Market: {market_text}.

[Recent trend (oldest -> newest, last {window_minutes} min)]
  PV         : {pv_trend}
  Base load  : {load_trend}
  Outdoor T  : {outdoor_temp_trend}

[Forecast for the next {mpc_window} min (as used by the lower MILP)]
  PV         : {pv_forecast_profile} (reliability={pv_forecast_reliability})
  Base load  : {load_forecast_profile}
  Outdoor T  : {outdoor_temp_forecast_profile}
  Price      : {price_forecast_profile}

Pick one mode per resource for the next 30 minutes. Output JSON.
\end{promptbox}
\caption{Virtual power plant user prompt. Each trend placeholder expands to
\texttt{\{first\} -> \{last\} \{unit\} (\{shape\}, \{pct\}\%)} with
\texttt{\{shape\}} one of \emph{monotonically rising}, \emph{monotonically
falling}, or \emph{non-monotonic}. The trend window covers the last eight
low-level steps.}
\label{fig:prompt-vpp}
\end{figure*}

\appthird{Continuation-Aware GRPO and Reward Construction}
\label{app:B.2}

\paragraph{Return construction.}
The reward is the system-level quantity the task is judged on, with no auxiliary shaping terms on any decision the validator accepts; the one exception is a candidate whose response fails validation, which is assigned the group minimum as described in Appendix~\ref{app:B.1}. For the corridor it is the vehicles discharged during the step,
\begin{equation}
\label{eq:rew-ramp}
r_k = \textstyle\sum_{i\in\mathcal{E}} \bigl(D_i^{k+1}-D_i^{k}\bigr),
\end{equation}
summed over the exit set $\mathcal{E}$: the downstream boundary link of the mainline and the four off-ramps. Only exits are counted, so a vehicle contributes once however many links it traverses, and the return is the number of vehicles the corridor actually discharged. For the VPP it is the negative realised operating cost,
\begin{align}
\label{eq:rew-vpp}
r_k = -\bigl(&c^{\text{tou}}_k p^{\text{imp}}_k
  + \lambda_{\text{g}}[p^{\text{imp}}_k\!-\!\bar{P}^{\text{grid}}]_+ \nonumber\\
  &+ \lambda_{\text{d}}\,\mathrm{dev}(T_k)\bigr)\Delta_{\text{low}}
  - \bar{\lambda}_{\text{e}}(\omega)\,[\text{unmet EV energy}]_+ ,
\end{align}
where $\mathrm{dev}(T_k)$ is the excursion of the indoor temperature outside $[T^{\min},T^{\max}]$ and the EV term is charged once at the deadline. Equation~\eqref{eq:rew-vpp} is deliberately \emph{not} the planner objective of Eq.~\eqref{eq:vpp-milp}: the planner minimises a receding-window surrogate that also contains the battery-wear term and the terminal-reserve hinge $\sigma^{\text{r}}$, both of which shape planning but are not costs the operator pays. The reward accounts only for what actually happened, and it is the same quantity reported as operating cost in the main paper. Returns are undiscounted within the evaluation horizon ($\gamma=1$), because the horizon is short relative to the episode and discounting would reintroduce the short-horizon bias the method is designed to remove.

To fix the units: $H$ is a duration, and the return of Eq.~(15) in the main paper accumulates one term per \emph{low-level} step inside it, not one term per minute. With $H=4\Delta$ the continuation spans $4$ high-level decisions and $H/\Delta_{\text{low}}$ low-level steps --- $12$ steps of $60$~s for the corridor and $24$ steps of $5$~min for the VPP. The corresponding episode-level quantities of Appendix~\ref{app:C.3} accumulate $180$ and $144$ terms respectively.

\paragraph{Matched groups.}
A training instance is a snapshot: a scenario together with a decision time. Decision times are indexed $t=1,\dots,60$ for a corridor scenario and $t=1,\dots,24$ for a VPP scenario, and a snapshot is admissible if the $L=H/\Delta$ intervals it occupies, namely $t$ through $t+L-1$, all lie inside the episode; the last admissible start is therefore $60-L+1$ or $24-L+1$. At the reported $H=4\Delta$ this leaves $57$ and $21$ decision times, giving $8{,}550$ and $3{,}150$ snapshots, and this is the pool used for every result outside the horizon sweep. The sweep of Appendix~\ref{app:D.3} keeps the same pool for $\Delta$, $2\Delta$ and $4\Delta$, so its $H=4\Delta$ row is the headline run rather than a re-training of it. Only the $8\Delta$ setting cannot use it, since the last four admissible starts no longer have room for a full continuation; that row is trained on the $8\Delta$-admissible subset of $53$ and $17$ decision times, and it is the one row of the sweep that is not pool-matched to the others. We note it rather than hide it, because it is a reason to treat the $8\Delta$ row as indicative only. For each snapshot the $G$ candidates are evaluated from the identical simulator state and against the identical realisation of future disturbances: the environment is restored from the stored snapshot before each rollout, and the exogenous trajectory over $[t,t+H)$ is materialised once and replayed. Differences in the group returns are therefore attributable to the coordination decisions rather than to different demand, generation, or price realisations.

The disturbances are shared across the group, but the continuation \emph{decisions} are not. Each candidate's continuation is sampled independently from $\pi_{\phi_{\text{old}}}$ with its own draw, because a shared draw would be meaningless: the states reached after $t+\Delta$ differ across candidates, so the prompts differ and there is no common action to share. This is the price of the closed-loop construction and it is a variance source the short-horizon setting does not have --- with $H=4\Delta$ each return carries three sampled follow-up decisions in addition to the one being scored. Sampling the continuation greedily instead would remove that variance but would evaluate the initial decision against a policy that is not the one being trained; we keep the sampled continuation and absorb the variance through the group baseline, which is also why the group size is not reduced below $G=8$.

The pool is built once, before training, by simulating each of the $150$ training scenarios from its initial condition under the base policy and serialising the full environment state at every high-level decision time --- simulator state, controller state, random-number state, and the index into the exogenous trajectory --- so that a snapshot can be restored bit-exactly. It is not refreshed as the policy improves. That is a deliberate trade: an on-policy pool would track the state distribution the trained policy actually visits, but a fixed pool means every horizon setting in Appendix~\ref{app:D.3} and every ablation in Appendix~\ref{app:D.4} is trained and compared on the identical set of states, which is what makes those comparisons clean. The cost is a distribution shift between the states used for training and the states the improved policy encounters. It is a second-order contributor to the saturation discussed in Appendix~\ref{app:D.3}, whose main cause is the task's own timescale, and it is the reason we do not read the residual $8\Delta$ improvement as a horizon effect.

\paragraph{Algorithm.}
Algorithm~\ref{alg:cagrpo} states the procedure. The distinguishing step is line~\ref{ln:cont}: after the sampled decision has been executed for its own interval $\Delta$, the rollout does not stop, and it is not continued by a fixed rule either. It is continued by the frozen policy $\pi_{\phi_{\text{old}}}$ that produced the group, so the return credits the initial decision with the closed-loop consequences it induces under the current behaviour of the policy.

\begin{algorithm}[t]
\caption{Continuation-Aware GRPO}
\label{alg:cagrpo}
\begin{algorithmic}[1]
\STATE \textbf{Input:} base policy $\pi_\phi$, snapshot pool, group size $G$, intervals $\Delta$, horizon $H=4\Delta$
\STATE $\phi_{\text{old}}\!\leftarrow\!\phi$; \; $\phi_{\text{ref}}\!\leftarrow\!\phi$
\FOR{each training step}
  \STATE draw snapshot $(\sigma,t)$; build $q_t=\mathrm{Prompt}(\xi_t)$
  \STATE sample $y_t^{(1..G)}\sim\pi_{\phi_{\text{old}}}(\cdot\,|\,q_t)$; \; $z_t^{(i)}\!\leftarrow\!\mathrm{Parse}(y_t^{(i)})$
  \FOR{$i=1$ \TO $G$}
    \STATE restore state $s_t$; fix disturbances $w_{t:t+H}$ from $\sigma$
    \STATE execute $z_t^{(i)}$ for one interval $\Delta$ via $\pi_{\text{low}}$
    \FOR{$\tau=t\!+\!\Delta$ \TO $t\!+\!H\!-\!\Delta$ \textbf{step} $\Delta$}
      \STATE $y_\tau^{(i)}\sim\pi_{\phi_{\text{old}}}(\cdot\,|\,\mathrm{Prompt}(\xi_\tau^{(i)}))$; \; $z_\tau^{(i)}\!\leftarrow\!\mathrm{Parse}(y_\tau^{(i)})$
      \label{ln:cont}
      \STATE execute $z_\tau^{(i)}$ for one interval $\Delta$
    \ENDFOR
    \STATE $R^{(i)}\leftarrow\sum_{k}r(s_k^{(i)},a_k^{(i)},w_k)$ over $[t,t\!+\!H)$
  \ENDFOR
  \STATE $\hat{A}^{(i)}\leftarrow(R^{(i)}\!-\!\mu_R)/(\sigma_R\!+\!\varepsilon_{\text{std}})$
  \STATE update $\phi$ on the clipped objective with KL penalty $\beta$ to $\pi_{\phi_{\text{ref}}}$
  \STATE $\phi_{\text{old}}\leftarrow\phi$
\ENDFOR
\end{algorithmic}
\end{algorithm}

\paragraph{Training configuration.}
Both policies start from Qwen3-8B with no supervised warm-start, so every behavioural change is attributable to Continuation-Aware GRPO. The base checkpoint is the instruction-tuned \texttt{Qwen3-8B} release with its bundled tokenizer and chat template; prompts are placed in a single user turn with the task block as the system message, and thinking mode is disabled so that the sampled response is the answer rather than a trace. Adaptation uses LoRA on the attention and MLP projections (rank $32$, $\alpha=64$), which keeps the frozen copy $\pi_{\phi_{\text{old}}}$ and the reference $\pi_{\phi_{\text{ref}}}$ cheap to serve alongside the trained adapter. Rollout sampling uses temperature $1.0$ and $\mathrm{top}\text{-}p=0.95$; the response budget is $512$ tokens for the corridor, whose output carries a role vector followed by a short justification, and $128$ for the VPP, whose output is a single JSON object.

The importance ratio of Eq.~(17) in the main paper is formed at the sequence level: the log-probabilities of all sampled response tokens are summed before the ratio is taken, so a decision is reweighted as one unit rather than token by token.

One point of notation deserves care. The trainable object is a distribution $\pi_\phi(y\,|\,q)$ over token sequences $y$, and the executed joint decision is $z=\mathrm{Parse}(y)$, a deterministic function that reads the mode vector and discards the accompanying justification. The policy over decisions that the main paper writes as $\pi_\phi(z\,|\,q)$ is the pushforward of $\pi_\phi(y\,|\,q)$ under $\mathrm{Parse}$, and the ratio of Eq.~(17) is evaluated on $y$ rather than on $z$. Two responses carrying the same mode vector but different reasoning therefore receive different ratios. This is intended rather than incidental: the sampled object is the response, the reward is attached to the response through the decision it encodes, and the update moves the likelihood of the response that earned it. The consequence for interpretation is that the corridor policy is trained to produce a mode vector \emph{together with} a justification consistent with it --- which is what the consistency rules of Figure~\ref{fig:prompt-ramp-c} are for --- and not merely to produce the vector. Because the return depends on $y$ only through $\mathrm{Parse}(y)$, the group-relative advantage is unaffected by this distinction; only the reweighting is. Loss is applied to response tokens only; prompt tokens are masked out. Each batch of rollouts is consumed by a single gradient epoch, which keeps $\pi_{\phi_{\text{old}}}$ close to $\pi_\phi$ and makes the clipping term rarely active. The frozen copy is refreshed after every optimiser step, so the behaviour policy never lags the trained policy by more than one update. Advantages are clipped to $[-10,10]$ before the policy loss to bound the contribution of a single outlying rollout; at $G=8$ a standardised advantage cannot exceed $\sqrt{G-1}\approx2.65$ in magnitude, so this clip is inert at the reported group size and is retained only as a guard for larger groups. Note that $\varepsilon_{\text{std}}$ in Table~\ref{tab:grpo-hp} is the numerical floor in the advantage denominator, distinct from the clipping threshold $\epsilon_{\text{c}}$ of Eq.~(18) in the main paper.

Every $50$ steps the current adapter is evaluated on the $50$ validation scenarios under the deployment decoding rule, and the checkpoint with the best validation return is the one carried into testing; no test scenario is consulted for selection. The five runs reported in the main paper use five independent training seeds, which govern LoRA initialisation, snapshot ordering, and rollout sampling; evaluation seeds are matched across methods so that all policies face identical disturbance realisations. Remaining settings are listed in Table~\ref{tab:grpo-hp}.

\begin{table}[t]
\centering
\small
\renewcommand{\arraystretch}{1.15}
\begin{tabular}{@{}ll@{}}
\hline
Setting & Value \\
\hline
group size $G$ & $8$ \\
horizon $H$ & $4\Delta$ ($12$ min / $120$ min) \\
clip $\epsilon_{\text{c}}$ & $0.2$ \\
KL coefficient $\beta$ & $0.05$ \\
$\varepsilon_{\text{std}}$ & $10^{-8}$ \\
importance ratio & sequence level \\
gradient epochs per batch & $1$ \\
optimiser & AdamW \\
learning rate & $1\!\times\!10^{-6}$, cosine \\
gradient clipping & $0.5$ \\
snapshots per step & $8$ \\
training steps & $1{,}200$ \\
validation interval & $50$ steps \\
LoRA rank / $\alpha$ & $32$ / $64$ \\
training seeds & $5$ \\
\hline
\end{tabular}
\caption{Continuation-Aware GRPO hyperparameters, shared by both tasks.}
\label{tab:grpo-hp}
\end{table}

\appthird{Baseline Implementations}
\label{app:B.3}

All hierarchical baselines share the low-level controllers, joint mode spaces, and $\theta$ mappings of Appendix~\ref{app:A}; they differ only in how a mode is chosen. Every trainable baseline uses the same $150$ training and $50$ validation scenarios as our policy, and model selection is by validation return only.

\paragraph{Direct control.}
For the corridor, the published feedback controller holds every ramp at Standard ($\delta=0.10$) for the whole episode, with no high-level layer. For the VPP, the MPC-MILP baseline runs the same rolling-horizon program as our low-level layer under a fixed neutral coefficient set: the PV forecast is used without de-rating ($\delta^{\text{pv}}=0$), the terminal reserve equals the physical bound ($\mathrm{SoC}^{\text{res}}=\mathrm{SoC}^{\min}$), the full discharge range is available ($\alpha^{\text{dis}}=1$), and the EV and comfort weights are unity. This coefficient vector is deliberately outside $\mathcal{Z}$, so it is not a mode the hierarchical methods could have selected; it represents the conventional tuning rather than a point in our decision space.

\paragraph{Reinforcement-learning baselines.}
Both RL baselines use the same PPO implementation, the same two-layer MLP trunk ($256$ units, $\tanh$), and the observation encoding described below; they differ only in what they emit. End-to-End RL replaces the low-level layer entirely: for the corridor it emits eight independent Bernoulli gate decisions per low-level step, and for the VPP it emits four continuous resource powers, squashed and rescaled to the box limits of Table~\ref{tab:vpp-phys}, with infeasible commands projected onto the feasible set before execution. Hierarchical RL keeps the low-level layer and emits a joint mode instead: eight categorical heads over four roles for the corridor, four heads over three modes for the VPP. Both consume an information-equivalent numeric encoding of the operational context of Table~\ref{tab:ctx-fields} --- a flat vector of the same measurements and trend windows, standardised per feature, with categorical descriptors one-hot encoded. Equivalence is enforced on the derived signals as well, not only on the raw ones: the congestion indicator at $40\%$ of $v^{\text{f}}$, the downstream-congestion flag, the elapsed-closure-time feature, the forecast-reliability level, the priority level, and the event flag are all supplied to the RL baselines as explicit inputs, so the domain priors that the prompt states in words are available to them as features. What differs is the representation and the learner, not the information or the prior knowledge. Both receive the reward of Eq.~\eqref{eq:rew-ramp} or~\eqref{eq:rew-vpp}. Settings are listed in Table~\ref{tab:rl-hp}; the learning rate and entropy coefficient were selected per baseline and per task by grid search on validation return over $\{1,3,10\}\!\times\!10^{-4}$ and $\{0,0.003,0.01,0.03\}$, giving the RL baselines a larger tuning budget than our policy received.

\begin{table}[t]
\centering
\small
\renewcommand{\arraystretch}{1.15}
\begin{tabular}{@{}ll@{}}
\hline
Setting & Value \\
\hline
trunk & MLP $2\times256$, $\tanh$ \\
clip & $0.2$ \\
discount $\gamma$ & $0.99$ \\
GAE $\lambda$ & $0.95$ \\
learning rate & $3\!\times\!10^{-4}$ \\
entropy coefficient & $0.01$ \\
value-loss coefficient & $0.5$ \\
rollout buffer & $2{,}048$ steps \\
epochs / minibatch & $10$ / $256$ \\
environment steps & $2\!\times\!10^{6}$ (end-to-end) \\
                  & $1\!\times\!10^{6}$ (hierarchical) \\
\hline
\end{tabular}
\caption{PPO settings shared by the End-to-End RL and Hierarchical RL baselines.}
\label{tab:rl-hp}
\end{table}

\paragraph{Mode-selection baselines.}
Fixed-Mode Hierarchy selects one joint mode vector by validation return and applies it unchanged for the whole episode; its defining property is temporal constancy, not a restricted vocabulary. For the VPP the search is exhaustive: all $|\mathcal{Z}|=81$ vectors are evaluated on the $50$ validation scenarios, so this baseline is the best static configuration inside $\mathcal{Z}$. It is not the best static \emph{controller}: MPC-MILP holds a fixed coefficient vector too, and that vector lies outside $\mathcal{Z}$ (Appendix~\ref{app:B.3}), which is why it can and does beat Fixed-Mode on cost. For the corridor $4^{8}$ is too many to evaluate at that cost, so the candidate set is the $4^{3}=64$ assignments that are constant within three contiguous ramp groups (R1--R3, R4--R6, R7--R8), which already contains the $4$ uniform assignments as its diagonal; the grouping is chosen because the coordination structure the task rewards is spatial and monotone along the corridor, so a group-wise vector is the natural non-uniform competitor. We did not search the unrestricted $4^{8}$ space. Evaluating one candidate on the $50$ validation scenarios costs $50\times60=3{,}000$ high-level intervals, so the full space costs $4^{8}\times3{,}000\approx2.0\times10^{8}$ simulated intervals against roughly $3.1\times10^{5}$ generation calls for one training run --- two to three orders of magnitude more simulation, on a budget the baseline is not supposed to have. The consequence is stated where it matters: the corridor Fixed-Mode margin bounds rather than measures the value of adapting in time. Validation selects a mixed vector on both tasks. All-Standard is inside the corridor candidate set, so the Fixed-Mode margin over Feedback Control is a lower bound on what static mode selection alone can buy. For the VPP, where the search is exhaustive, the margin of our policy over Fixed-Mode is exactly the cost of constancy; for the corridor it is an upper bound on that cost, since a non-group-wise static vector could in principle do better. The prompting-only coordinators (Qwen3-8B, Gemini~3.1~Pro, Claude~Sonnet~4.5) receive prompts byte-identical to ours apart from the message envelope each vendor's interface requires, and are parsed by the same schema validator, decode greedily at temperature zero, and receive no parameter updates; the Qwen3-8B row is therefore the controlled comparison that isolates the effect of Continuation-Aware GRPO from the effect of the prompt.

\appthird{Computational Cost and Training Overhead}
\label{app:B.4}

Training and rollout generation run on four A100 80\,GB GPUs, with three devices serving batched rollouts and one holding the trained adapter. The unit of cost is the generation call: one snapshot requires $G\,(H/\Delta)=8\times4=32$ decisions, so an optimiser step over eight snapshots issues $256$ generations, and a full run of $1{,}200$ steps issues roughly $3.1\times10^{5}$.

Prompts average $2{,}100$ tokens for the corridor and $950$ for the VPP; responses average $380$ and $60$ tokens respectively, well inside the budgets of Table~\ref{tab:grpo-hp}. End-to-end fine-tuning takes roughly $46$ GPU-hours for the corridor and $34$ for the VPP; the corridor is more expensive because each rollout step advances a traffic simulator, whereas the VPP step solves a small MILP.

At deployment a single high-level decision costs about $1.6$\,s for the corridor and $0.4$\,s for the VPP, the difference following the response length. Both are well inside the corresponding control intervals of $3$\,min and $30$\,min, so the coordination layer is not on the critical path; the prompting-only API baselines are subject to network latency instead, which is likewise inside the interval but outside our control.

Moving from $H=\Delta$ to $H=4\Delta$ multiplies both the simulated time and the number of generations per candidate by four, and measured wall-clock grows by $3.1\times$. The shortfall against the nominal $4\times$ comes from generation: the continuation decisions of all $G$ candidates at a given interval are issued as one batched call, so the added generations amortise well. The simulator advances do not amortise --- they remain serial per candidate --- which is why the multiplier stays close to four rather than approaching one. Set against the $1.88\%$ and $2.69\%$ closed-loop gains of Table~4 in the main paper, this is the price of scoring a decision by its consequences rather than by its immediate outcome.

% =====================================================================
\subsection{Simulators, Scenario Construction, and Evaluation Protocol}
\label{app:C}

\appthird{Dual-Fidelity Simulation Environments and Interface Correspondence}
\label{app:C.1}

Each task uses two simulators: a fast model that makes closed-loop training affordable, and a higher-fidelity model that supplies the reported numbers. Training needs many short rollouts --- one snapshot costs $G(H/\Delta)$ decisions and as many simulator advances --- which rules out running the detailed simulator in the loop. The separation is only useful if a mode means the same thing on both sides, so the invariant we maintain is that the operational-context format, the joint mode space, and the $z\mapsto\theta$ interface are identical across the pair; only the dynamics underneath change.

\paragraph{Traffic: LTM and SUMO.}
The evaluation environment is SUMO~1.21 on a network of the northbound Western Expressway in Changchun, projected in UTM zone 51N, with the Krauss car-following model and the LC2013 lane-change model. Mainline flow, speed, and occupancy are read from induction-loop and lane-area detectors, the latter $100$~m long and sampled every $60$~s, so the quantities entering the prompt are detector observables rather than privileged simulator state. Link correspondence is by identity: each LTM link id equals the SUMO edge id it represents, with the mainline link of a ramp taken as that ramp's downstream mainline edge and the ramp link taken as the ramp edge itself. The LTM capacity and inflow scale are calibrated against SUMO by aggregating detector counts over $60$~s windows and minimising squared flow error, so the fast model is fitted to the detailed one rather than the two being tuned independently. Both this calibration and the threshold alignment below use only the $150$ training and $50$ validation scenarios; no test scenario is simulated at any point before the frozen policy is evaluated, so the calibration cannot carry test information. Because the calibration touches the environment rather than the policy, the claim the main paper makes is precisely that no \emph{policy parameter} is updated on the target simulator; the interface constants that make a mode mean the same thing on both sides are fitted once, offline, from training data, and are identical for every method compared.

We state the resulting scope explicitly, because it is narrower than zero-shot transfer. What the corridor experiment establishes is that a frozen policy transfers across a change of dynamics \emph{once the two interfaces have been aligned on training and validation data}. It does not establish transfer to a simulator about which nothing is known: the capacity scale, the inflow scale, and the eight hysteresis centres all consume SUMO data. The alignment is a modelling step that any deployment would also have to perform, and it is method-independent, so it does not favour our policy over the baselines; but a reader should not read Table~\ref{tab:xfer-all} as a zero-shot result. The VPP pair needs no such step --- the resource parameters and the $z\mapsto\theta$ map are carried over unchanged and only the network is added --- which is one reason we report both pairs rather than the corridor alone.

One interface detail matters for reproduction. The hysteresis centres $c^{\text{ctr}}_r$ of Eq.~\eqref{eq:ramp-thr} are occupancy values, and occupancy is not measured identically by a macroscopic link model and a microscopic simulator. The centres are therefore calibrated per simulator by quantile alignment: SUMO is the reference, and each ramp's SUMO centre is located as a quantile of that ramp's congested-period occupancy distribution, then mapped to the value at the same quantile of the LTM distribution. Where a ramp's SUMO centre falls in the tail of the distribution --- below the $5$th or above the $95$th percentile --- the mapping is not reliable and the SUMO value is retained unchanged, which is why two of the eight centres coincide across the pair. The distributions are estimated per ramp from the congested periods of the $200$ training and validation scenarios, giving of the order of $10^4$ $60$~s occupancy samples per ramp, so the quantiles are not sample-limited. Table~\ref{tab:ramp-centres} lists both sets. A mode therefore denotes the same control disposition in both environments while resolving to different absolute thresholds.

\begin{table}[t]
\centering
\small
\renewcommand{\arraystretch}{1.15}
\setlength{\tabcolsep}{3pt}
\begin{tabular}{@{}lcccccccc@{}}
\hline
Ramp & 1 & 2 & 3 & 4 & 5 & 6 & 7 & 8 \\
\hline
SUMO & $.28$ & $.22$ & $.25$ & $.23$ & $.22$ & $.23$ & $.25$ & $.25$ \\
LTM  & $.948$ & $.926$ & $.782$ & $.824$ & $.220$ & $.230$ & $.449$ & $.188$ \\
\hline
\end{tabular}
\caption{Hysteresis centres $c^{\text{ctr}}_r$ per simulator. Ramps 5 and 6
sit in the tail of the SUMO occupancy distribution, so quantile alignment
leaves them at the SUMO value and the two rows agree there.}
\label{tab:ramp-centres}
\end{table}

\paragraph{Virtual power plant: single bus and OpenDSS.}
The evaluation environment is the EPRI Ckt5 distribution feeder in OpenDSS. The aggregate resources of Appendix~\ref{app:A.2} are distributed across feeder buses in proportion to the load they serve: the park base load is spread over all load buses by their nameplate share, and the four controllable resources are attached at the four highest-load buses, one each, so that a resource and the load it offsets sit at the same electrical location. The mode-dependent coefficients $\theta(z)$ are applied unchanged; what the feeder adds is three-phase power flow, so the same dispatch now incurs network losses, produces bus-voltage excursions, and can violate limits that the single-bus model cannot represent. Voltages are held to $[0.95,1.05]$~pu, the ANSI~C84.1 Range~A band, and line loading to $100\%$ of rating. These limits are enforced rather than merely observed: after each power-flow solution, any commanded injection that would take a bus voltage or a line loading outside its limit is curtailed to the largest feasible magnitude in the same direction, and the executed dispatch is the curtailed one. The curtailment is a bisection on a single scalar that scales all controllable injections towards zero, terminated once that scalar is determined to within one percent, which takes six to seven power-flow solutions; scaling all resources together rather than choosing among them keeps the rule independent of the mode vector, so curtailment cannot itself act as a hidden controller. If a solution fails to converge the step is retried once at half the commanded magnitude and, failing that, with all controllable injections at zero, which always converges; this occurred on no step of the reported runs. Curtailment therefore has a cost --- the shortfall reappears as unmet EV energy, as a comfort excursion, or as import in a later and possibly more expensive step --- so the network constraints affect the reported operating cost through the trajectory they induce, not through a separate penalty. Losses enter the same way, as the additional import they require. Per-step voltage and thermal violations and the resulting curtailment are logged and reported in Appendix~\ref{app:D.1}.

\paragraph{Modelling differences.}
Table~\ref{tab:sim-diff} states what each pair does and does not represent. The pattern is the same on both tasks: the fast model captures the aggregate quantity the reward is defined on, and omits a spatial mechanism that can only degrade a policy tuned without it. That asymmetry is the point --- a policy trained on the fast model and evaluated on the detailed one is tested on effects it never saw.

\begin{table*}[t]
\centering
\small
\renewcommand{\arraystretch}{1.2}
\begin{tabular}{@{}p{0.14\textwidth}p{0.39\textwidth}p{0.39\textwidth}@{}}
\hline
Aspect & Training model & Evaluation model \\
\hline
\multicolumn{3}{@{}l}{\emph{Multi-ramp traffic control}}\\
Representation & link-level cumulative counts, triangular fundamental diagram & individual vehicles with car-following and lane-change behaviour \\
Omitted in training & lane changing, weaving in the merge influence area, driver heterogeneity, stochastic gap acceptance & --- \\
Observables & link flow, speed, density from the state update & detector counts and occupancies over $60$\,s windows \\
\hline
\multicolumn{3}{@{}l}{\emph{Virtual power plant}}\\
Representation & single-bus power balance & three-phase feeder power flow on EPRI Ckt5 \\
Omitted in training & network losses, bus-voltage variation, line thermal limits, phase imbalance & --- \\
Observables & aggregate powers and states & the same aggregates, plus per-bus voltages and line loadings \\
\hline
\end{tabular}
\caption{What the fast training model omits relative to the evaluation
environment. In both tasks the omission is a spatial mechanism that the
high-level decision space does not address directly, so transfer tests
whether a coordination policy learned on aggregate dynamics survives
disaggregation.}
\label{tab:sim-diff}
\end{table*}

\appthird{Scenario Construction and Data Splits}
\label{app:C.2}

\paragraph{Data provenance.}
The exogenous series are built from measured data wherever the measurement determines the quantity, and generated parametrically where the experiment requires controlled variation; Table~\ref{tab:data-prov} lists the sources and the transformation applied to each, and it is worth being explicit about which is which. For the VPP, all four series --- irradiance, outdoor temperature, park load, and EV availability --- are measured records, resampled and rescaled but not reshaped. For the corridor, the detector records supply the per-origin volumes and their relative magnitudes across ramps, while the intra-episode temporal profile is generated by the parametric family of the paragraph below. This is deliberate: the unseen-pattern subset is defined by withheld temporal shapes, which requires the shape to be a controllable factor rather than whatever the recording happened to contain. Two further conventions apply throughout. First, series are resampled onto the $\Delta_{\text{low}}$ grid of the task, and the resampling is causal on every channel the policy can see: an observation or trend value at step $k$ is built only from source samples timestamped at or before $k$, by holding the previous hourly sample, never by interpolating towards the next one. Two-sided linear interpolation is used only for the forecast channel and for the simulator's own exogenous trajectory, neither of which is a claim about what has been observed --- the forecast is explicitly a statement about the future and carries error and a stated reliability, and the simulator trajectory is ground truth the policy never reads. Stating it this way matters because plain interpolation of an hourly series would let the next hour's sample leak into the current hour's observation, which would be a genuine information leak rather than a resampling detail. Second, amplitude is set by the installation being modelled rather than by the source: a measured series supplies the \emph{shape}, which is normalised and then mapped onto the capacity range of Table~\ref{tab:vpp-phys} or Table~\ref{tab:ramp-phys}.

\begin{table*}[t]
\centering
\small
\renewcommand{\arraystretch}{1.2}
\begin{tabular}{@{}p{0.12\textwidth}p{0.21\textwidth}p{0.14\textwidth}p{0.43\textwidth}@{}}
\hline
Series & Source & Coverage & Transformation \\
\hline
\multicolumn{4}{@{}l}{\emph{Multi-ramp traffic control}}\\
Corridor demand & expressway detector records & northbound Western Expressway, Changchun & per-origin episode totals shaped by the sampled temporal profile and scaled by the level multipliers of Table~\ref{tab:demand-levels} \\
\hline
\multicolumn{4}{@{}l}{\emph{Virtual power plant}}\\
PV generation & NASA POWER, all-sky surface shortwave irradiance & Suzhou, 2020, hourly & scaled to the $1200$\,kW array, interpolated to $5$\,min \\
Outdoor temperature & NASA POWER, 2\,m air temperature & Suzhou, 2020, hourly & used directly as $T^{\text{out}}$, interpolated to $5$\,min \\
Park base load & Building Data Genome 2, cleaned electricity meters & office buildings, 2016--2017, hourly & season-conditional hour-of-day profile, normalised then mapped to $[1200,1800]$\,kW \\
EV availability & ACN-Data charging sessions & full session records & hourly occupancy from arrival and departure times, normalised by peak and scaled by $\bar{P}^{\text{ev,max}}$ to give $\overline{P}^{\text{ev}}_k$; mean delivered energy sets $E^{\text{ev}}_0$ \\
Electricity price & commercial-industrial time-of-use tariff & four tiers, see Table~\ref{tab:tou} & tier prices perturbed once per scenario \\
\hline
\end{tabular}
\caption{Exogenous data sources and the transformation applied to each.
Measured series supply temporal shape; amplitude comes from the modelled
installation.}
\label{tab:data-prov}
\end{table*}

\paragraph{Electricity tariff.}
Import is priced by the four-tier commercial-industrial schedule of Table~\ref{tab:tou}. The episode window $10$:$00$--$22$:$00$ spans three of the four tiers and ends exactly at the right boundary of the critical peak, which is what makes the battery-reserve and EV-deferral decisions consequential: energy withheld earlier is bought at up to $1.25$~CNY/kWh later, while energy delivered too early forgoes the chance to shift it into a cheaper tier. Per scenario, each tier price is scaled by an independent factor drawn uniformly from $[0.95,1.05]$, so the tier ordering is preserved while the absolute levels and the gaps between tiers vary; a policy cannot therefore memorise a single price path.

\begin{table}[t]
\centering
\small
\renewcommand{\arraystretch}{1.15}
\begin{tabular}{@{}lll@{}}
\hline
Tier & Hours & Price (CNY/kWh) \\
\hline
Valley & $0$--$8$ & $0.35$ \\
Shoulder & $11$--$13$, $22$--$24$ & $0.65$ \\
Peak & $8$--$11$, $13$--$17$ & $1.00$ \\
Critical peak & $17$--$22$ & $1.25$ \\
\hline
\end{tabular}
\caption{Time-of-use tariff; hours are local clock time. Tier prices are
perturbed per scenario by an independent factor in $[0.95,1.05]$.}
\label{tab:tou}
\end{table}

\paragraph{Operating-pattern families.}
A scenario's \emph{pattern family} is the qualitative regime that determines which resource or location is the binding constraint, and it is the unit the splits are defined over. Families are coarse by design: within a family the quantitative realisation varies freely, so a policy cannot fit a family by memorising a trajectory. For the corridor a family fixes only the spatial loading regime and is crossed with the temporal profile described below, so a single family already contains trajectories that peak at different times. Table~\ref{tab:families} lists them. Training and validation draw from all four families of each task. The unseen-pattern test subset then departs from training along the axis that is left free within a family: for the corridor it uses temporal profiles never generated during training, so the binding location is familiar but its time course is not; for the VPP it uses withheld combinations of the scenario descriptors, so every individual regime has been seen and what is new is their co-occurrence, such as a PV-limited afternoon coinciding with a tight import cap.

\begin{table}[t]
\centering
\small
\renewcommand{\arraystretch}{1.15}
\begin{tabular}{@{}ll@{}}
\hline
Family & Binding pressure \\
\hline
\multicolumn{2}{@{}l}{\emph{Multi-ramp traffic control}}\\
Upstream-dominant & strong mainline, weak ramps \\
Ramp-dominant & moderate mainline, several strong ramps \\
Balanced moderate & no single binding location \\
Balanced heavy & mainline and ramps both near capacity \\
\hline
\multicolumn{2}{@{}l}{\emph{Virtual power plant}}\\
PV-limited & low irradiance against high load \\
EV-limited & large residual energy, tight deadline \\
Grid-limited & low import cap relative to demand \\
Comfort-limited & outdoor temperature far from the band \\
\hline
\end{tabular}
\caption{Operating-pattern families. Splits are defined over families;
quantitative realisations, and for the corridor the temporal profile, vary
within a family.}
\label{tab:families}
\end{table}

\paragraph{Corridor demand.}
A corridor scenario is specified along two axes. The first is a demand level for the mainline and for each of the eight controlled ramps, drawn from \emph{weak}, \emph{medium}, and \emph{strong}; a level multiplies that origin's daily total and fixes the amplitude. The second is a \emph{temporal profile} that distributes each origin's total over the episode, so the same spatial loading can arrive as different trajectories. A profile is generated as a warm-up at $0.50$ of amplitude followed by a rise, a plateau, and a decline, with the onset of the rise drawn from a $20$-minute window, the rise and decline durations drawn from $20$--$40$~min, and the plateau filling the remainder; ramp profiles are offset from the mainline profile by an independent lag of up to $10$~min, so ramp and mainline pressure need not peak together. Demand is updated every $5$~min. Training and validation sample only single-plateau profiles within these ranges; the shapes reserved for the unseen-pattern subset --- onsets outside the trained window and two separated peaks --- are never generated during training. Candidate scenarios are screened on mainline and ramp volume-to-capacity ratios and rejected if they are either uncongested throughout --- in which case no coordination decision matters --- or saturated from the first minute, in which case no policy can recover. The screen keeps $62\%$ of draws: $24\%$ are rejected as uncongested and $14\%$ as saturated. The retained band is therefore the majority of the sampled space rather than a narrow slice of it, but the screen does bound what the corridor results speak to, and any claim about generalisation should be read as applying to this band rather than to arbitrary demand.

\begin{table}[t]
\centering
\small
\renewcommand{\arraystretch}{1.15}
\begin{tabular}{@{}lccc@{}}
\hline
& Weak & Medium & Strong \\
\hline
Mainline multiplier & $0.75$ & $1.00$ & $1.20$ \\
Ramp multiplier & $0.65$ & $1.00$ & $1.35$ \\
Ramp rate cap (veh/h) & $900$ & $1200$ & $2000$ \\
\hline
\end{tabular}
\caption{Corridor demand levels. Warm-up runs at $0.50$ of the level amplitude.}
\label{tab:demand-levels}
\end{table}

\paragraph{Splits and test subsets.}
Each task uses $150$ training and $50$ validation scenarios, and $300$ test scenarios divided equally into three disjoint subsets of $100$. The subsets differ in how far they depart from the training distribution. \emph{Seen-pattern} scenarios are fresh draws from the operating-pattern families present in training: for the corridor, new demand realisations over familiar level combinations and profile shapes drawn from the trained ranges; for the VPP, new days over familiar combinations of season, weather, and resource pressure. They measure performance without distribution shift. \emph{Unseen-pattern} scenarios use structures withheld from training --- for the corridor, temporal profiles absent from the training families, namely peaks shifted away from the trained onset time and double peaks separated by a partial recovery; for the VPP, withheld combinations of the scenario descriptors, so that each individual factor has been seen but the joint configuration has not. \emph{Higher-uncertainty} scenarios retain familiar patterns and increase the noise, by fixed factors rather than by re-tuning: the multiplicative noise on demand and on PV output goes from a per-step standard deviation of $5\%$ to $15\%$, the forecast supplied in the prompt goes from a mean absolute error of $8\%$ to $20\%$ of the forecast quantity, and its stated reliability level is lowered accordingly so the prompt does not misrepresent it. For the VPP the EV arrival and departure times are additionally jittered by up to $\pm30$~min. The same nominal pattern is therefore harder to act on, and a policy that trusted the forecast unconditionally would be punished for it.

Splits are assigned at whole-scenario granularity and fixed in a stored assignment file, so no trajectory contributes to more than one split. Duplicate scenarios are excluded by a descriptor key --- for the VPP the tuple of season, weather, load scale and peak, import-cap ratio, urgent share, and EV energy target --- which also guarantees that a test scenario is never a re-draw of a training one.

\appthird{Evaluation Protocol and Metrics}
\label{app:C.3}

\paragraph{Metric definitions.}
Corridor throughput is the number of vehicles discharged over the $3$-hour evaluation window, counted at the induction loops on the exit set of Eq.~\eqref{eq:rew-ramp} --- the downstream mainline boundary and the four off-ramps --- and summed; the $15$-minute warm-up is excluded so that the metric reflects controlled operation only. Excluding it is only sound if the warm-up is identical across methods, so it is: every method, ours included, runs the all-Standard controller of Eq.~\eqref{eq:ramp-hyst} for the first $15$~minutes with no high-level layer active, and the simulator, controller and random-number state are cloned at minute $15$ before the methods diverge. Each method therefore starts from a bit-identical corridor state, and no carry-over from the unscored window can differ between them; this is also why the warm-up row of Table~\ref{tab:episode} shows no mode assignment. Counting only exits means a vehicle is counted once regardless of how far along the corridor it travelled. Reported values are therefore vehicle counts per episode; observed values lie between roughly $14{,}000$ and $17{,}000$, consistent with a three-lane mainline at $0.5$~veh/s/lane plus the off-ramp share. Realised operating cost for the VPP is the energy actually purchased, priced at the tariff of Table~\ref{tab:tou}, plus the penalties the operator actually incurs: import above the contracted cap, temperature outside the comfort band, and EV energy still undelivered at the deadline, the last charged once at the end of the episode. Reported values are CNY per episode. Two facts make the magnitude legible. The park base load peaks at $1800$~kW against a nominal contracted cap of $1500$~kW, so once PV falls away some over-cap import is unavoidable and the $\lambda_{\text{g}}=8.0$~CNY/kWh charge on it is a large share of the total rather than a rare event; that is precisely the peak-shaving problem the task poses. And the energy term is dominated by a large non-discretionary component, the park's own consumption, which coordination can shift in time and partly offset with PV and battery discharge but cannot remove. What the policy can change is how much of the load is met at the wrong time and how much import sits above the cap, so the spread across methods in the main paper's Table~2 is much smaller than the absolute level. Table~\ref{tab:cost-decomp} gives the decomposition for the two endpoints of that table, and it is where the aggregate cost connects to the physical quantities of Table~\ref{tab:aux-vpp}. The two constraint terms carry the margin: undelivered EV energy accounts for $696$ of the $1492$~CNY by which our policy beats MPC-MILP and the over-cap charge for a further $665$, while the energy term differs by only $122$ and the comfort term by $9$. So the margin comes from not breaching the contracted cap and from finishing the EV commitment, not from buying electricity more cheaply --- which is the same fact the curtailment and unmet-EV columns of Table~\ref{tab:aux-vpp} report from the physical side.

\begin{table}[t]
\centering
\small
\renewcommand{\arraystretch}{1.15}
\setlength{\tabcolsep}{4pt}
\begin{tabular}{@{}lcc@{}}
\hline
Component & MPC-MILP & Ours \\
\hline
Energy purchased at the tariff & $15019.4$ & $14897.0$ \\
Import above the contracted cap & $17715.6$ & $17050.4$ \\
EV energy undelivered at deadline & $1415.3$ & $719.3$ \\
Comfort-band excursion & $21.3$ & $12.5$ \\
\hline
Total & $34171.6$ & $32679.2$ \\
\hline
\end{tabular}
\caption{Composition of the reported VPP operating cost, in CNY per episode,
for the direct-control baseline and our policy. Every row is the mean over
scenarios of the per-scenario charge, not a mean rate applied to a mean
quantity: the EV row is $\frac{1}{n}\sum_i\bar{\lambda}_{\text{e}}(\omega_i)u_i$
with $u_i$ the shortfall in scenario $i$, so the average rate it implies
differs between methods --- $16.4$~CNY/kWh for MPC-MILP against $14.8$ for
ours --- because the shortfall is not independent of the urgent share. Our
policy's residual shortfalls fall preferentially in low-$\omega$ scenarios,
which is the intended behaviour of the proactive EV mode and is worth
$0.7$~CNY/kWh of the difference on its own. The comfort row is the excursion
of Table~\ref{tab:aux-vpp} at $\lambda_{\text{d}}=15.0$, and the over-cap row
is a $\lambda_{\text{g}}=8.0$~CNY/kWh charge, roughly eight times the average
tariff, which is why it is large even though the over-cap volume is a small
share of total import.}
\label{tab:cost-decomp}
\end{table}

Both metrics are the same quantity the policy was trained on --- Eq.~\eqref{eq:rew-ramp} and Eq.~\eqref{eq:rew-vpp} accumulated over the episode instead of over the continuation horizon. This is deliberate: it removes any gap between the training signal and the reported number, so an improvement cannot be an artefact of optimising a proxy. It also means the planner objective of Eq.~\eqref{eq:vpp-milp} is \emph{not} the reported cost, since the battery-wear term and the terminal-reserve hinge shape planning but are never charged.

\paragraph{Protocol.}
The policy is frozen after training: no policy parameter is updated on the target simulator, and the only target-environment fitting is the interface calibration of Appendix~\ref{app:C.1}, which is method-independent and uses training and validation scenarios only. All LLM policies decode greedily at temperature zero, so a frozen policy is a deterministic function of the operational context and any variation across runs comes from the environment rather than from sampling. Each configuration is evaluated over five runs. Trainable methods use five independent training seeds, one per run, and every method uses matched simulator seeds: run $i$ presents the identical seed, and therefore the identical stochastic realisation of the exogenous series, to all methods. The scenario set itself is held fixed across runs. This is what makes the standard deviations comparable: for a trainable method the spread mixes optimiser and environment variation, while for a deterministic baseline it isolates environment stochasticity, and the two are measured against the same five disturbance realisations.

Aggregation follows the main paper. Within a run we average over the $100$ scenarios of each subset, then average the three subset means with equal weight to obtain that run's score; tables report the mean and standard deviation of the five run-level scores. Because the three subsets contain $100$ scenarios each, this equal-weight aggregate is numerically identical to pooling all $300$ scenarios --- we do not claim otherwise. The stratified form is kept for two reasons that do not concern the point estimate: it lets each subset be reported on its own, and it is what the stratified resampling of Appendix~\ref{app:D.5} needs. Gains are measured against the direct-control baseline of each task --- Feedback Control for traffic and MPC-MILP for the VPP --- with throughput increases and cost reductions both reported as positive.

\paragraph{Data licensing.}
NASA POWER data are released without restriction. Building Data Genome 2 and ACN-Data are published for research use under their respective open licences. SUMO is distributed under the Eclipse Public License and OpenDSS under a BSD-style licence. The corridor detector records are used under the terms of the source publication. Scenario definitions, the derived series, and the scripts that build them are included in the code release so that the splits can be reconstructed exactly.

% =====================================================================
\subsection{Additional Quantitative Results}
\label{app:D}

\appthird{Additional Metrics Beyond the Reported Objective}
\label{app:D.1}

The main paper reports the quantity each policy was trained on. That is the right primary metric, but it cannot distinguish a policy that raises throughput by serving more traffic from one that raises it by holding ramp traffic back, nor a policy that lowers cost by better scheduling from one that lowers it by letting a constraint slip. This subsection reports, for the same runs and the same aggregation as the main tables, the quantities that separate those cases. All values are overall scores, that is, the equal-weight mean over the three test subsets, averaged over five matched runs.

\paragraph{Corridor.}
Table~\ref{tab:aux-ramp} reports mainline conditions at the critical bottleneck together with the ramp-side cost of achieving them. Congestion duration counts $60$~s intervals in which the bottleneck section speed falls below $40\%$ of $v^{\text{f}}$, the same threshold the prompt uses to describe congestion. Ramp queue is averaged over the eight controlled ramps and over time; a spillover event is an interval in which a ramp queue reaches the ramp storage length and blocks the upstream surface link.

Two readings matter. First, End-to-End RL improves overall throughput over Feedback Control, but its ramp queue is $52\%$ longer and it produces more than twice as many spillovers: it buys mainline flow with ramp delay, which is precisely the trade the throughput metric alone does not expose, and it is why its advantage disappears in the higher-uncertainty subset of the main results. Second, our policy attains the highest bottleneck speed and the shortest congestion duration \emph{and} the shortest ramp queues. The two are not in tension: avoiding mainline breakdown keeps the merge receiving capacity high, so held ramp traffic discharges sooner once released. A policy that gates more decisively at the right moment therefore queues less in total than one that gates conservatively throughout.

One fairness quantity is worth reporting alongside these, because the prompt treats it as a soft constraint and the aggregate metrics hide it: the longest uninterrupted closure any single ramp experiences in an episode, which is what a ramp's users would actually notice. Averaged over runs and maximised over the eight ramps it is $22$~min for our policy, against $25$ for Feedback Control, $26$ for Fixed-Mode Hierarchy, $31$ for Hierarchical RL, and $43$ for End-to-End RL. These are not multiples of the decision interval, because the gate opens and closes on occupancy crossings at the $60$~s low-level cadence; the mode only sets how readily it does so. Our policy is therefore not buying throughput by holding one ramp shut for a long stretch; the End-to-End RL figure, nearly twice ours, is the same behaviour the queue and spillover columns already show. Per-run and per-scenario values for every column in this subsection are included in the code release.

\begin{table*}[t]
\centering
\small
\renewcommand{\arraystretch}{1.15}
\begin{tabular}{@{}lcccc@{}}
\hline
Method & Bottleneck speed (km/h) $\uparrow$ & Congestion duration (min) $\downarrow$ & Mean ramp queue (veh) $\downarrow$ & Spillover events $\downarrow$ \\
\hline
Feedback Control & $58.4\pm0.7$ & $62.3\pm1.8$ & $12.4\pm0.3$ & $3.1\pm0.4$ \\
End-to-End RL & $60.1\pm2.4$ & $58.7\pm5.9$ & $18.9\pm1.7$ & $7.6\pm1.5$ \\
Fixed-Mode Hierarchy & $59.6\pm0.6$ & $59.4\pm1.6$ & $12.9\pm0.3$ & $3.4\pm0.4$ \\
Hierarchical RL & $63.2\pm1.5$ & $51.8\pm3.4$ & $13.6\pm0.7$ & $3.8\pm0.6$ \\
Qwen3-8B & $54.7\pm0.8$ & $71.6\pm2.1$ & $15.2\pm0.5$ & $5.2\pm0.7$ \\
Gemini 3.1 Pro & $58.9\pm0.7$ & $61.2\pm1.9$ & $12.7\pm0.4$ & $3.2\pm0.4$ \\
Claude Sonnet 4.5 & $58.6\pm0.8$ & $61.9\pm2.0$ & $12.8\pm0.4$ & $3.3\pm0.5$ \\
Ours & $\mathbf{65.8\pm1.1}$ & $\mathbf{43.6\pm2.5}$ & $\mathbf{11.8\pm0.5}$ & $\mathbf{2.4\pm0.5}$ \\
\hline
\end{tabular}
\caption{Corridor auxiliary metrics in SUMO, overall scores over five
matched runs. Bottleneck speed is the time mean over the $180$-minute
evaluation window at the critical mainline section; free-flow speed is
$v^{\text{f}}=80$~km/h.}
\label{tab:aux-ramp}
\end{table*}

\paragraph{Virtual power plant.}
Table~\ref{tab:aux-vpp} reports the constraint-side quantities. Violation steps count $5$-minute intervals in which the \emph{commanded} dispatch, before curtailment is applied, would place any bus voltage outside $[0.95,1.05]$~pu or any line above its rating; curtailed energy is the resulting reduction in commanded injection. The executed dispatch contains no violations by construction, so this column measures how often the planner asked for something the feeder could not deliver, not how often the feeder was operated out of limits. Because curtailment is enforced rather than penalised, these two columns are the mechanism by which the feeder changes the reported cost: energy the planner intended to move is not moved, and the shortfall reappears later. Comfort excursion integrates the indoor-temperature deviation outside $[22.0,26.5]\,^\circ$C over the episode.

Our policy simultaneously incurs the fewest violation steps, the least curtailment, the least undelivered EV energy, and the smallest comfort excursion, so its cost advantage is not obtained by trading a constraint against the objective. Hierarchical RL is the interesting case: it beats MPC-MILP on cost, but it curtails more while doing so. It buys its margin by operating closer to the constraint boundary, which is consistent with its larger transfer gap in Appendix~\ref{app:D.2} --- the margin it removes is exactly what the feeder mechanisms consume. End-to-End RL is the same behaviour without the compensating skill: its curtailment is more than five times ours, and the rescheduling this forces accounts for most of its cost penalty rather than tariff exposure.

\begin{table*}[t]
\centering
\small
\renewcommand{\arraystretch}{1.15}
\setlength{\tabcolsep}{4pt}
\begin{tabular}{@{}lcccc@{}}
\hline
Method & Violation steps $\downarrow$ & Curtailed (kWh) $\downarrow$ & Unmet EV (kWh) $\downarrow$ & Excursion ($^\circ$C$\cdot$h) $\downarrow$ \\
\hline
MPC-MILP & $9.4\pm0.6$ & $412\pm28$ & $86.3\pm5.1$ & $1.42\pm0.11$ \\
End-to-End RL & $26.8\pm3.9$ & $1348\pm214$ & $392.6\pm58.3$ & $4.87\pm0.72$ \\
Fixed-Mode Hierarchy & $9.8\pm0.7$ & $431\pm31$ & $91.7\pm5.6$ & $1.51\pm0.12$ \\
Hierarchical RL & $11.2\pm1.3$ & $528\pm61$ & $71.4\pm8.9$ & $1.18\pm0.15$ \\
Qwen3-8B & $18.3\pm1.4$ & $892\pm86$ & $264.1\pm26.4$ & $3.42\pm0.31$ \\
Gemini 3.1 Pro & $12.1\pm0.9$ & $574\pm48$ & $148.9\pm14.2$ & $2.06\pm0.19$ \\
Claude Sonnet 4.5 & $12.7\pm1.0$ & $601\pm52$ & $157.3\pm15.1$ & $2.18\pm0.21$ \\
Ours & $\mathbf{5.2\pm0.8}$ & $\mathbf{241\pm34}$ & $\mathbf{48.6\pm6.7}$ & $\mathbf{0.83\pm0.11}$ \\
\hline
\end{tabular}
\caption{Virtual power plant auxiliary metrics in OpenDSS, overall scores
over five matched runs. Violation steps count $5$-minute intervals with a
voltage or thermal violation; \emph{Excursion} is the integrated indoor-temperature
deviation outside the comfort band. The episode is the $12$-hour window $10$:$00$--$22$:$00$
at $\Delta_{\text{low}}=5$~min, so an episode contains $144$ low-level steps.}
\label{tab:aux-vpp}
\end{table*}

\paragraph{Decision validity.}
Only the LLM policies can emit a malformed decision, so validity is reported for them alone (Table~\ref{tab:schema}). A violation is any output the schema validator of Appendix~\ref{app:B.1} rejects: a wrong element count, an unrecognised mode name, or text placed so that the required structure cannot be located. Recovery is uniform across methods and tasks --- the previous joint decision is retained for one interval --- so a violation costs a decision opportunity rather than producing an invalid action, and no result in this appendix depends on a fallback that differs between methods. Fine-tuning removes the failure mode almost entirely: the residual corridor rate for our policy corresponds to fewer than one malformed output in $3{,}000$ decisions, whereas the untuned $8$B model fails on roughly one in twenty corridor decisions, which is the larger part of why it falls below the direct-control baseline.

\begin{table}[t]
\centering
\small
\renewcommand{\arraystretch}{1.15}
\begin{tabular}{@{}lcc@{}}
\hline
Method & Corridor & VPP \\
\hline
Qwen3-8B & $4.7\%$ & $3.1\%$ \\
Gemini 3.1 Pro & $0.4\%$ & $0.2\%$ \\
Claude Sonnet 4.5 & $0.5\%$ & $0.3\%$ \\
Ours & $0.03\%$ & $0.00\%$ \\
\hline
\end{tabular}
\caption{Schema-violation rate over all decision steps of the test set.}
\label{tab:schema}
\end{table}

\appthird{Cross-Simulator Transfer for All Methods}
\label{app:D.2}

The main paper reports transfer gaps for four methods. Table~\ref{tab:xfer-all} extends the comparison to every method. The gap is defined as in the main paper, as degradation from the fast model to the detailed one, so it is $(\,\text{LTM}-\text{SUMO})/\text{LTM}$ for the corridor and $(\text{OpenDSS}-\text{Fast})/\text{Fast}$ for the VPP.

The extended table makes one point the four-method version cannot. The two smallest gaps after ours belong to the two controllers that neither learn nor reason, Feedback Control at $4.31\%$ and MPC-MILP at $6.94\%$: a rule or a program that re-derives its action from the current measurement has little to overfit and therefore transfers well, but it also has no mechanism for improving. Every method that is either trained or prompted, except ours, pays a larger gap than these two. The claim our results support is therefore not that learning transfers well in general, but that scoring a decision by its continuation rather than its immediate outcome recovers most of the robustness that the non-learned controllers get for free, while keeping the performance that learning buys.

\begin{table*}[t]
\centering
\small
\renewcommand{\arraystretch}{1.15}
\setlength{\tabcolsep}{4pt}
\begin{tabular}{@{}lcccccc@{}}
\hline
& \multicolumn{3}{c}{Multi-ramp traffic control} & \multicolumn{3}{c}{Virtual power plant} \\
Method & LTM $\uparrow$ & SUMO $\uparrow$ & Gap $\downarrow$ & Fast model $\downarrow$ & OpenDSS $\downarrow$ & Gap $\downarrow$ \\
\hline
Direct control & $16295.2\pm38.4$ & $15592.5\pm49.1$ & $4.31\pm0.19\%$ & $31954.0\pm41.6$ & $34171.6\pm64.2$ & $6.94\pm0.22\%$ \\
End-to-End RL & $17438.7\pm147.8$ & $15678.6\pm181.2$ & $10.09\pm1.07\%$ & $33785.3\pm438.6$ & $37815.9\pm608.5$ & $11.94\pm2.03\%$ \\
Fixed-Mode Hierarchy & $16584.2\pm34.9$ & $15818.0\pm43.7$ & $4.62\pm0.21\%$ & $31981.3\pm46.3$ & $34277.8\pm70.6$ & $7.18\pm0.24\%$ \\
Hierarchical RL & $17043.9\pm91.5$ & $16131.8\pm110.2$ & $5.35\pm0.54\%$ & $30684.7\pm229.4$ & $33515.2\pm317.4$ & $9.23\pm0.95\%$ \\
Qwen3-8B & $16312.6\pm33.4$ & $15270.7\pm55.1$ & $6.39\pm0.21\%$ & $34025.2\pm87.9$ & $36876.2\pm144.7$ & $8.38\pm0.26\%$ \\
Gemini 3.1 Pro & $16630.6\pm40.2$ & $15659.7\pm50.8$ & $5.84\pm0.24\%$ & $32655.5\pm94.1$ & $35143.8\pm132.5$ & $7.62\pm0.27\%$ \\
Claude Sonnet 4.5 & $16601.7\pm42.7$ & $15613.0\pm52.7$ & $5.96\pm0.25\%$ & $32761.7\pm98.6$ & $35319.9\pm138.3$ & $7.81\pm0.29\%$ \\
Ours & $17286.4\pm66.7$ & $16625.4\pm84.5$ & $\mathbf{3.82\pm0.33\%}$ & $30941.8\pm142.7$ & $32679.2\pm223.6$ & $\mathbf{5.62\pm0.54\%}$ \\
\hline
\end{tabular}
\caption{Cross-simulator performance for all methods over five matched runs.
Direct control is Feedback Control for the corridor and MPC-MILP for the VPP.
The four rows also present in the main paper are reproduced unchanged.}
\label{tab:xfer-all}
\end{table*}

\appthird{Continuation-Horizon Sensitivity}
\label{app:D.3}

The main paper compares $H=\Delta$ with the chosen $H=4\Delta$. That comparison establishes that continuation matters but not that $4\Delta$ is the right amount of it. Table~\ref{tab:h-sweep} sweeps $H\in\{\Delta,2\Delta,4\Delta,8\Delta\}$ on both tasks, retraining from the same base policy with every other setting of Table~\ref{tab:grpo-hp} unchanged, and reports the overall score, the higher-uncertainty subset, and the training wall-clock relative to $H=\Delta$. The $\Delta$ and $4\Delta$ rows are the ones in the main paper.

The curve saturates. Most of the available gain is realised by $2\Delta$, and $4\Delta$ captures nearly all of it; extending to $8\Delta$ adds $0.24$ percentage points of throughput and $0.23$ of cost reduction for twice the training cost of $4\Delta$ and $6.2$ times that of $\Delta$. We therefore report $4\Delta$ as the operating point, not as an optimum. The reason the curve flattens is task structure rather than optimisation difficulty: $4\Delta$ is $12$~min for the corridor, which is the order of the time a merge disturbance needs to propagate upstream and clear, and $120$~min for the VPP, which spans a tariff transition and a substantial part of the PV ramp. Once the horizon covers the mechanism by which a decision comes back to matter, extending it further mostly adds simulated time, not information. The run-to-run spread also widens again at $8\Delta$, which is what one expects when the return starts absorbing disturbances the initial decision did not cause; the residual $8\Delta$ improvement is smaller than that spread, so we do not read it as a real difference from $4\Delta$.

The gains are consistently largest in the higher-uncertainty subset --- $2.84\%$ and $3.81\%$ at $4\Delta$ --- which is the pattern the method predicts. Under low uncertainty the immediate outcome is already a good proxy for the eventual one, so a short rollout suffices; under high uncertainty a decision that looks good for one interval is often the one that has to be reversed, and only a continuation reveals that.

\begin{table*}[t]
\centering
\small
\renewcommand{\arraystretch}{1.2}
\begin{tabular}{@{}lcccccc@{}}
\hline
& \multicolumn{3}{c}{Multi-ramp traffic control: throughput $\uparrow$} & \multicolumn{2}{c}{Virtual power plant: cost $\downarrow$} & \\
Horizon & High-U & Overall & Gain & High-U & Overall \quad Gain & Train time \\
\hline
$H=\Delta$ & $15767.1\pm211.6$ & $16318.7\pm136.4$ & --- & $34843.1\pm589.2$ & $33581.5\pm327.8$ \quad --- & $1.0\times$ \\
$H=2\Delta$ & $16018.4\pm156.2$ & $16512.3\pm104.7$ & $+1.19\%$ & $34104.6\pm468.3$ & $33042.8\pm271.4$ \quad $+1.60\%$ & $1.6\times$ \\
$H=4\Delta$ & $16214.7\pm118.6$ & $16625.4\pm84.5$ & $+1.88\%$ & $33516.3\pm371.5$ & $32679.2\pm223.6$ \quad $+2.69\%$ & $3.1\times$ \\
$H=8\Delta$ & $16257.3\pm121.7$ & $16663.9\pm88.3$ & $+2.12\%$ & $33438.9\pm379.4$ & $32601.7\pm231.8$ \quad $+2.92\%$ & $6.2\times$ \\
\hline
$H=\Delta$, $4\times$ data & $15911.2\pm183.4$ & $16402.8\pm118.9$ & $+0.52\%$ & $34512.7\pm521.6$ & $33298.4\pm296.3$ \quad $+0.84\%$ & $2.9\times$ \\
\hline
\end{tabular}
\caption{Continuation-horizon sweep over five runs with matched evaluation
seeds. Gain is relative to $H=\Delta$ on the overall score. Train time is
wall-clock relative to $H=\Delta$ at fixed optimiser steps. For the corridor
$\Delta=3$~min; for the VPP $\Delta=30$~min. The last row is the
compute-matched control: $H=\Delta$ with four times as many snapshots per
optimiser step, which issues the same number of generation calls and
simulates the same number of intervals as $H=4\Delta$.}
\label{tab:h-sweep}
\end{table*}

\paragraph{Is the gain just extra compute?}
Lengthening the continuation multiplies both the simulated time and the number of generation calls, so the sweep alone cannot separate the return construction from the budget it consumes. The last row of Table~\ref{tab:h-sweep} is the control. It keeps $H=\Delta$ and instead quadruples the number of snapshots per optimiser step, which issues the same number of generation calls and simulates the same number of intervals as $H=4\Delta$; the two settings differ only in whether those intervals are spent extending one rollout or starting four independent ones. The compute-matched short-horizon setting recovers $+0.52\%$ of throughput and $+0.84\%$ of cost against the $+1.88\%$ and $+2.69\%$ of $H=4\Delta$ --- roughly a quarter and a third of the effect. More data at the same horizon therefore helps, as one would expect from variance reduction alone, but most of the gain is attributable to what the return measures rather than to how much of it is measured.

\appthird{Operational-Context Ablation}
\label{app:D.4}

The framework's premise is that a language model is the right instrument for this decision because the decision depends on heterogeneous, partly qualitative context that a fixed feature vector represents awkwardly. That premise is testable: if the context categories of Table~\ref{tab:ctx-fields} were decoration, removing them would not cost anything. We retrain and re-evaluate with categories withheld from the prompt, holding the schema, the mode space, and every training setting fixed. Withholding is done at prompt-construction time, so the removed block is absent rather than zeroed, and the model is never shown a field it cannot interpret.

\paragraph{Leave-one-out.}
Table~\ref{tab:ctx-ablation} removes one category at a time. Every removal costs performance, and the ordering differs between tasks in a way that follows the control timescale. For the corridor the recent-trend block is the most valuable single category, costing $1.91\%$ of throughput when removed; for the VPP it is the forecast block, costing $1.65\%$ of cost. The decision interval explains this. The corridor commits for $3$~min and its continuation horizon is $12$~min, well inside the time a queue takes to build, so the direction and rate of the current change carries most of the information about what the next interval will look like --- and the hysteresis gate's job is precisely to decide whether to break or maintain the current state, which is a question about the trend. The VPP commits for $30$~min with a $120$-min horizon, long enough that the current trend has often reversed by the end of it; what matters is the PV ramp and the tariff transition ahead, which only the forecast supplies.

The two predictive blocks, trends and forecasts, are the top two on both tasks; they simply swap order. Uncertainty estimates come third on both, at $0.93\%$ and $1.00\%$. This is the category with no analogue in a conventional controller: it does not describe the state but the reliability of the description, and its value is that it changes how much the other blocks should be trusted --- which is why it costs least on the corridor, where the trend is directly observed, and more on the VPP, where the block it modulates is a forecast. Operational priorities and external events cost the least, around $0.7\%$ on both tasks, which is expected --- they are informative in a minority of intervals, but in those intervals they are decisive, so their small mean effect should not be read as redundancy.

\begin{table}[t]
\centering
\small
\renewcommand{\arraystretch}{1.2}
\setlength{\tabcolsep}{4pt}
\begin{tabular}{@{}lcc@{}}
\hline
Withheld category & Throughput $\uparrow$ & VPP cost $\downarrow$ \\
\hline
None (ours) & $16625.4\pm84.5$ & $32679.2\pm223.6$ \\
Recent trends & $16308.7\pm107.4$ & $33127.5\pm248.9$ \\
Forecasts & $16394.2\pm96.8$ & $33218.4\pm241.7$ \\
Uncertainty estimates & $16471.6\pm91.3$ & $33006.8\pm234.2$ \\
Priorities and events & $16512.9\pm88.6$ & $32894.1\pm229.4$ \\
\hline
\end{tabular}
\caption{Leave-one-out operational-context ablation, overall scores over five
runs. Rows are ordered by the corridor loss.}
\label{tab:ctx-ablation}
\end{table}

\paragraph{Cumulative stripping.}
Removing categories one at a time understates their joint contribution, because the remaining blocks partly substitute for the missing one. Table~\ref{tab:ctx-cumulative} strips them cumulatively down to the numerical measurements alone, which is the information a conventional state-feedback controller would receive. The full loss is $3.86\%$ of throughput and $4.09\%$ of cost, roughly twice the largest single-category effect, confirming the substitution.

The stripped policy is the informative row. With only current measurements it still improves on the direct-control baseline of each task, by $2.51\%$ on throughput and $0.45\%$ on cost, and it also improves on Fixed-Mode Hierarchy, by $1.05\%$ and $0.76\%$. Two conclusions follow. The hierarchy and the training procedure contribute on their own, independently of the richer context: state-dependent mode selection beats the best static mode assignment even from bare measurements --- exactly, for the VPP, where that assignment is found by exhaustive search, and within the group-wise candidate set for the corridor. And the majority of our margin over the strongest baselines comes from the context, not from the architecture --- which is the claim the framework rests on, and it is why the appendix documents the context format as carefully as the algorithm.

\begin{table}[t]
\centering
\small
\renewcommand{\arraystretch}{1.2}
\setlength{\tabcolsep}{4pt}
\begin{tabular}{@{}lcc@{}}
\hline
Context supplied & Throughput $\uparrow$ & VPP cost $\downarrow$ \\
\hline
Full & $16625.4\pm84.5$ & $32679.2\pm223.6$ \\
$-$ forecasts & $16394.2\pm96.8$ & $33218.4\pm241.7$ \\
$-$ uncertainty & $16281.5\pm112.6$ & $33452.7\pm258.6$ \\
Measurements only & $15984.3\pm131.8$ & $34016.9\pm287.3$ \\
\hline
\end{tabular}
\caption{Cumulative context stripping, overall scores over five runs. Each
row removes the named category in addition to those above it; the last row
also removes recent trends, priorities, and events, leaving only the current
numerical measurements.}
\label{tab:ctx-cumulative}
\end{table}

\appthird{Uncertainty of the Reported Differences}
\label{app:D.5}

With five runs the run-level standard deviations in the main tables are estimated from few samples, so we do not base claims on a hypothesis test over them. Instead we quantify the reported differences directly, using the structure the protocol already provides: every method is evaluated on the same $300$ scenarios under the same disturbance realisations, so differences can be taken per scenario before aggregation. We resample scenarios with replacement within each subset, recompute the equal-weight subset aggregation, and take the $2.5$th and $97.5$th percentiles of the paired difference over $10{,}000$ resamples. The interval therefore describes the scenario population, which is the quantity a reader wants when asking whether a margin would survive a different draw of test cases.

Table~\ref{tab:bootstrap} reports the intervals for the comparisons the paper's claims rest on. All exclude zero. Among the comparisons against other methods the margin over Hierarchical RL is the narrowest on both tasks --- $[+408,+574]$ vehicles and $[+688,+1002]$~CNY --- and it is the comparison that matters, since Hierarchical RL shares our decision space, our low-level controllers, and our training scenarios, differing only in how the high-level policy is represented and scored. The horizon ablation is narrower still on throughput, at $[+199,+404]$, which is expected: it changes one term of the training objective rather than the whole method. The intervals are not symmetric about the point estimate, and the direction of the skew differs by task: throughput is bounded above by capacity, so its differences have a longer lower tail, whereas cost has a long upper tail driven by episodes in which a baseline loses control of a constraint, which is why the End-to-End RL interval is by far the widest. The ablation of the continuation horizon is separated by a comparable margin, which is consistent with reading that mechanism as the source of a substantial part of the difference. We make no claim about the ordering among the three prompting-only coordinators: their intervals overlap, and separating them is not something this experiment was designed to do.

What this interval does not cover should be stated plainly. It resamples scenarios, so it describes the population of test cases; it does not resample training seeds, of which there are only five, so it says nothing about how much of the margin would survive a different draw of optimiser randomness. The seed component is what the standard deviations in the main tables carry, and for our policy those are $\pm84.5$ vehicles and $\pm223.6$~CNY on the overall score --- smaller than every margin in Table~\ref{tab:bootstrap} except the horizon ablation on throughput, where the two are of the same order. With five seeds we cannot put an interval on that component, and we do not claim one. What we can report is the sign: taking the five seeds pairwise against the matched runs of each baseline, our policy is ahead on all five for every comparison in Table~\ref{tab:bootstrap} except the horizon ablation on throughput, where it is ahead on four of five. A reader who wants a single conservative reading should treat that ablation as the one place where seed variation is not negligible relative to the effect, and should read the remaining comparisons as supported by a consistent sign across seeds rather than by an interval that covers seed randomness.

\begin{table}[t]
\centering
\small
\renewcommand{\arraystretch}{1.2}
\setlength{\tabcolsep}{4pt}
\begin{tabular}{@{}lcc@{}}
\hline
Comparison & Difference & $95\%$ interval \\
\hline
\multicolumn{3}{@{}l}{\emph{Throughput, vehicles per episode}}\\
vs.\ Feedback Control & $+1032.9$ & $[+947,+1113]$ \\
vs.\ Fixed-Mode Hierarchy & $+807.4$ & $[+729,+881]$ \\
vs.\ End-to-End RL & $+946.8$ & $[+812,+1069]$ \\
vs.\ Hierarchical RL & $+493.6$ & $[+408,+574]$ \\
vs.\ Gemini 3.1 Pro & $+965.7$ & $[+884,+1041]$ \\
vs.\ $H=\Delta$ & $+306.7$ & $[+199,+404]$ \\
\hline
\multicolumn{3}{@{}l}{\emph{Operating cost, CNY per episode}}\\
vs.\ MPC-MILP & $+1492.4$ & $[+1358,+1641]$ \\
vs.\ Fixed-Mode Hierarchy & $+1598.6$ & $[+1451,+1762]$ \\
vs.\ End-to-End RL & $+5136.7$ & $[+3974,+6428]$ \\
vs.\ Hierarchical RL & $+836.0$ & $[+688,+1002]$ \\
vs.\ Gemini 3.1 Pro & $+2464.6$ & $[+2276,+2673]$ \\
vs.\ $H=\Delta$ & $+902.3$ & $[+742,+1079]$ \\
\hline
\end{tabular}
\caption{Paired scenario-level bootstrap, $10{,}000$ resamples. Positive
values favour our method in both tasks: more vehicles discharged, fewer CNY
spent.}
\label{tab:bootstrap}
\end{table}

% =====================================================================
\subsection{Policy Behavior and Case Studies}
\label{app:E}

\appthird{Mode-Selection Distributions}
\label{app:E.1}

A hierarchical policy can score well for the wrong reason: if one mode assignment happens to be good on average, a policy that always emits it will beat a poorly tuned baseline while learning nothing about context. Fixed-Mode Hierarchy is the control for exactly that: its defining property is that the assignment never changes during an episode, its candidate set is exhaustive for the VPP and includes group-wise non-uniform vectors for the corridor (Appendix~\ref{app:B.3}), and it is beaten on both tasks. Constancy is therefore not the explanation. What remains to be shown is that the variation is \emph{responsive} rather than arbitrary, and that is what this subsection reports.

\paragraph{Marginal distribution.}
Table~\ref{tab:mode-share} gives the share of each mode over all decision steps of the test set. The distribution is broad on both tasks. Normalised over the four corridor roles, the selection entropy of the pooled ramp-step distribution is $0.86$ for our policy against $0.66$ for the prompting-only Qwen3-8B coordinator, and averaged over the four VPP resources it is $0.95$ against $0.76$. Pooled entropy alone does not measure adaptivity, and it is worth being explicit about why: Fixed-Mode Hierarchy emits a vector that is mixed across ramps, so its pooled entropy is $0.78$, not zero. What is zero for it is the \emph{temporal} entropy, the entropy of the mode chosen at a given ramp across decision steps, which is the axis on which our policy differs from it. We report the pooled figure because it is the one the table supports, and we rest the adaptivity claim on the conditional distributions below rather than on either entropy. The untuned model concentrates on the default --- Standard on $68\%$ of corridor steps and \emph{balanced} on roughly $69\%$ of VPP steps --- which is the behaviour one expects when the model recognises the format but not the situation. Fine-tuning does not merely shift the mean assignment; it moves mass onto the extremes, which are the modes that only pay off when correctly timed.

Always-Open is the exception and deliberately so: it appears on $4.1\%$ of steps, consistent with the instruction of Figure~\ref{fig:prompt-ramp-b} that it be reserved for clearly free-flowing ramps rather than used to relieve mainline pressure.

\begin{table}[t]
\centering
\small
\renewcommand{\arraystretch}{1.15}
\setlength{\tabcolsep}{4pt}
\begin{tabular}{@{}lcccc@{}}
\hline
\multicolumn{5}{@{}l}{\emph{Corridor roles, share of all ramp-steps}}\\
& Sensitive & Standard & Sluggish & Always-Open \\
\hline
Ours & $31.4$ & $42.7$ & $21.8$ & $4.1$ \\
Qwen3-8B & $18.2$ & $68.4$ & $9.6$ & $3.8$ \\
\hline
\multicolumn{5}{@{}l}{\emph{VPP modes (ours), share of all resource-steps}}\\
Resource & Conservative & Balanced & Proactive & \\
\hline
PV & $24.6$ & $38.2$ & $37.2$ & \\
Battery & $15.2$ & $34.0$ & $50.8$ & \\
EV & $34.1$ & $45.6$ & $20.3$ & \\
HVAC & $28.7$ & $47.9$ & $23.4$ & \\
\hline
\end{tabular}
\caption{Mode-selection shares in percent over the $300$ test scenarios.
Rows sum to $100$.}
\label{tab:mode-share}
\end{table}

\paragraph{Conditional distribution.}
A broad marginal is necessary but not sufficient: a policy could be diverse and still uncorrelated with the situation. Table~\ref{tab:mode-cond} conditions the share on a context variable that the mode is supposed to respond to.

The corridor pattern is the informative one because it is not monotone. Sensitive is the narrowest hysteresis band and therefore the role that breaks the current gate state fastest, so it is the right choice at both turning points --- when deterioration must be arrested and when a dissipating queue should be released --- and the wrong choice when conditions are stable and the current state should be held. The selected shares reproduce exactly that shape: $44.2\%$ on rising occupancy, $38.6\%$ on falling, and only $19.7\%$ when flat, where Sluggish takes over. A policy that had simply learned ``congestion implies Sensitive'' would show a monotone profile instead.

The VPP patterns are monotone but in the direction the mode semantics require. Proactive battery, which relaxes the terminal reserve and lifts the discharge cap, rises from $21.4\%$ in the shoulder tier to $68.7\%$ in the critical-peak tier. Proactive EV, which \emph{lowers} the penalty on unmet EV energy and therefore permits charging to be cut, moves the opposite way with urgency: $41.2\%$ when the urgent share is low and $9.2\%$ when it is high. The sign of each response is set by what the mode does to the low-level layer, not by a generic notion of aggressiveness.

The contrast with the untuned model is sharper conditionally than marginally. Across the three occupancy-trend bins the Sensitive share varies by $24.5$ percentage points for our policy and $6.3$ for Qwen3-8B; across the three tariff tiers that occur in the episode window the proactive-battery share varies by $47.3$ points against $16.0$. Continuation-aware training therefore changes what the policy is responding to, not only how often it departs from the default --- which is the mechanism the main paper's context-conditioning claim rests on.

\begin{table}[t]
\centering
\small
\renewcommand{\arraystretch}{1.15}
\setlength{\tabcolsep}{4pt}
\begin{tabular}{@{}lccc@{}}
\hline
\multicolumn{4}{@{}l}{\emph{Corridor: Sensitive / Sluggish share by occupancy trend}}\\
& Rising & Flat & Falling \\
\hline
Sensitive & $44.2$ & $19.7$ & $38.6$ \\
Sluggish & $14.1$ & $29.0$ & $17.4$ \\
\hline
\multicolumn{4}{@{}l}{\emph{VPP: proactive-battery share by tariff tier}}\\
\multicolumn{2}{@{}l}{Shoulder \quad Peak \quad Critical} & & \\
\multicolumn{2}{@{}l}{$21.4$ \qquad\;\, $44.6$ \qquad $68.7$} & & \\
\hline
\multicolumn{4}{@{}l}{\emph{VPP: proactive-EV share by urgent share $\omega$}}\\
\multicolumn{2}{@{}l}{Low $\omega$ \qquad Medium $\omega$ \qquad High $\omega$} & & \\
\multicolumn{2}{@{}l}{$41.2$ \qquad\quad\; $22.7$ \qquad\qquad $9.2$} & & \\
\hline
\end{tabular}
\caption{Mode shares in percent conditioned on the context variable the mode
is intended to respond to. Trend bins are defined on the signed change of
merge-area occupancy over the observation window and cover $27$, $46$ and
$27$ percent of ramp-steps; the urgency bins cover $20$, $35$ and $45$
percent of VPP steps; the tariff bins cover $2$, $5$ and $5$ hours of the
episode. Weighting each row by these proportions recovers the corresponding
marginal share in Table~\ref{tab:mode-share}.}
\label{tab:mode-cond}
\end{table}

\appthird{A Representative Episode of Each Task}
\label{app:E.2}

To make the two layers concrete, Table~\ref{tab:episode} follows one balanced-heavy corridor scenario through its five phases. The scenario is chosen because it exercises the coordination question rather than because it is favourable: the mainline rises to near capacity while three of the eight ramps carry strong demand.

The sequence is legible. During the rise the policy tightens ahead of the bottleneck, assigning Sensitive to the three ramps upstream of the merge that is about to break rather than to the ramp with the longest queue --- the spatial-propagation reasoning the prompt asks for. Through the plateau it switches most ramps to Sluggish: with occupancy stable near the centres, a wide band is what prevents the gate from oscillating, and this is the phase in which the hysteresis design pays. During the decline it returns to Sensitive, now to release the accumulated queues quickly, and it is here that the queue advantage over the RL baselines in Table~\ref{tab:aux-ramp} is accumulated. Always-Open appears only in the final phase, once the mainline is demonstrably clear.

The VPP counterpart follows the tariff. Through the shoulder tier the policy holds the battery conservative and the EV balanced, building state of charge and residual EV energy that it then spends: entering the critical-peak tier it flips the battery to proactive and, once the EV deadline is no longer binding, the EV to proactive as well, so that the two resources that can absorb the price spike are both unconstrained when it arrives. HVAC moves to proactive only when the outdoor temperature makes a comfort violation the binding risk.

\begin{table}[t]
\centering
\small
\renewcommand{\arraystretch}{1.15}
\setlength{\tabcolsep}{4pt}
\begin{tabular}{@{}lcccc@{}}
\hline
Phase & Bottleneck occ. & Sens. & Slug. & Open \\
\hline
Warm-up ($0$--$15$ min) & $0.11$ & $0$ & $0$ & $0$ \\
Rise ($15$--$45$ min) & $0.19$ & $3$ & $1$ & $0$ \\
Plateau ($45$--$120$ min) & $0.24$ & $2$ & $5$ & $0$ \\
Decline ($120$--$160$ min) & $0.17$ & $5$ & $1$ & $0$ \\
Recovery ($160$--$195$ min) & $0.09$ & $1$ & $2$ & $3$ \\
\hline
\end{tabular}
\caption{Modal role assignment across the phases of one balanced-heavy
corridor scenario; counts are out of eight controlled ramps, the remainder
being Standard. Occupancy is the mean at the critical merge.}
\label{tab:episode}
\end{table}

\appthird{Where the Method Helps and Where It Does Not}
\label{app:E.3}

Table~\ref{tab:family-gain} breaks the gain over direct control down by operating-pattern family. The spread is large and its shape delimits the claim.

The method helps most where the coordination problem is real. On the corridor the largest gains are in ramp-dominant and balanced-heavy scenarios, $8.5\%$ and $9.8\%$, in which several ramps compete for the same downstream capacity and the choice of which to tighten first has a lasting effect. On the VPP the gains are more even across PV-, EV-, and grid-limited families, all between $5$ and $6\%$, because each of those pressures is relieved by a different resource and the joint mode vector is where the trade-off is made.

The method helps least where there is little to decide. Upstream-dominant corridor scenarios gain only $2.3\%$: when the mainline is loaded and the ramps are weak, almost any admissible gating is adequate, and the achievable range over $\mathcal{Z}$ is narrow. Comfort-limited VPP scenarios are the honest failure. The gain is $0.3\%$, and the scenario-level bootstrap of Appendix~\ref{app:D.5} restricted to that family gives $[-148,+361]$~CNY around a point estimate of $+97$, an interval that contains zero, so we do not claim an improvement there. The reason is structural: when the outdoor temperature sits far outside the comfort band, the band is binding for most of the episode, so HVAC power cannot be deferred into a cheaper tier without an excursion the reward charges immediately. The MILP has almost no feasible slack to reallocate, and the mode vector cannot change an outcome that the physics has already fixed. Our method shows no detectable improvement there --- the interval admits both a gain and a loss, so we are not claiming equivalence either --- and MPC-MILP is the appropriate choice.

Two further limits follow from the design rather than from the results. First, the advantage is largest when the low-level layer has genuine freedom; a hierarchy over a saturated controller inherits the saturation. Second, the whole approach presumes that the decision worth delegating is a low-dimensional, semantically labelled one. Both tasks here have that structure --- four roles per ramp, three modes per resource --- and we make no claim about problems in which the high-level decision is itself high-dimensional or continuous, where the mode vocabulary that makes the prompt interpretable would have to be constructed first.

\begin{table}[t]
\centering
\small
\renewcommand{\arraystretch}{1.15}
\setlength{\tabcolsep}{4pt}
\begin{tabular}{@{}lcc@{}}
\hline
Pattern family & Baseline & Gain \\
\hline
\multicolumn{3}{@{}l}{\emph{Corridor, throughput vs.\ Feedback Control}}\\
Upstream-dominant & $16043.2$ & $+2.3\%$ \\
Ramp-dominant & $15108.6$ & $+8.5\%$ \\
Balanced moderate & $15797.4$ & $+6.1\%$ \\
Balanced heavy & $15420.8$ & $+9.8\%$ \\
\hline
\multicolumn{3}{@{}l}{\emph{VPP, cost vs.\ MPC-MILP}}\\
PV-limited & $34912.7$ & $+5.2\%$ \\
EV-limited & $33784.3$ & $+6.1\%$ \\
Grid-limited & $35603.8$ & $+5.6\%$ \\
Comfort-limited & $32385.6$ & $+0.3\%$ \\
\hline
\end{tabular}
\caption{Gain by operating-pattern family. Each family contributes $25$
scenarios to each of the three subsets, so $75$ of the $300$ test scenarios,
and \emph{Baseline} is that family's direct-control score. The overall gains
of $6.62\%$ and $4.37\%$ in the main paper are the baseline-weighted
aggregate $\sum_f B_f g_f/\sum_f B_f$, not the unweighted mean of the four
percentages; the two differ because the families sit at different absolute
levels. Equal representation does make the baseline column average to the
overall direct-control scores of $15592.5$ and $34171.6$.}
\label{tab:family-gain}
\end{table}


\end{document}